\documentclass[journal]{./templates/IEEEtran}

\newcommand{\pdftitle}{Energy Aware Deep Reinforcement Learning  Scheduling for Sensors Correlated in Time and Space}

\usepackage[nolist,nohyperlinks]{acronym}
\usepackage{amsmath}
\usepackage{amssymb} 
\usepackage{breqn}
\usepackage{balance}
\usepackage{cite}
\usepackage{framed}
\usepackage{siunitx}
\usepackage{soul}
\usepackage{multirow}
\usepackage{etoolbox}
\usepackage{standalone}
\usepackage{textcomp}
\usepackage{multirow,booktabs}
\usepackage{amsmath}
\usepackage{amssymb}
\usepackage{bm}
\usepackage{mathtools} 
\usepackage{optidef} 

\usepackage{xcolor, colortbl, collcell}
\colorlet{activecolour}{blue}

\usepackage{graphicx}
\usepackage{subcaption}
\usepackage[font=small]{caption}

\graphicspath{{./images/}}

\usepackage{tikz}
\usepackage{pgfplots}
\usepgfplotslibrary{groupplots}

\usepackage{url}
\usepackage[bookmarks=false]{hyperref}
\hypersetup{
  backref,
  hidelinks,
  colorlinks=false,
  breaklinks=true,
  bookmarksopen=false,
  pdftitle=\pdftitle,
  pdfauthor={Author}
}

\newtoggle{authorcomment}
\toggletrue{authorcomment}

\definecolor{LightCyan}{rgb}{0.88,1,1}
\definecolor{Gray}{gray}{0.85}

\usetikzlibrary{arrows,decorations.pathmorphing,backgrounds,positioning,fit,petri}
\usetikzlibrary{shapes.misc}
\tikzset{cross/.style={cross out, draw=black, minimum size=2*(#1-\pgflinewidth), inner sep=0pt, outer sep=0pt},
	cross/.default={1pt}}
\usetikzlibrary{patterns}

\begin{document}

\newcommand\copyrighttext{%
  \footnotesize \textcopyright 2021 IEEE. Personal use of this material is permitted.
  Permission from IEEE must be obtained for all other uses, in any current or future
  media, including reprinting/republishing this material for advertising or promotional
  purposes, creating new collective works, for resale or redistribution to servers or
  lists, or reuse of any copyrighted component of this work in other works.
  DOI: \href{https://doi.org/10.1109/JIOT.2021.3114102}{10.1109/JIOT.2021.3114102}}
\newcommand\copyrightnotice{%
\begin{tikzpicture}[remember picture,overlay]
\node[anchor=south,yshift=0pt] at (current page.south) {\fbox{\parbox{\dimexpr\textwidth-\fboxsep-\fboxrule\relax}{\copyrighttext}}};
\end{tikzpicture}%
}

\bstctlcite{IEEEexample:BSTcontrol}

\title{\pdftitle}

\author{\IEEEauthorblockN{Jernej Hribar, Andrei Marinescu, Alessandro Chiumento, and Luiz A. DaSilva}

\thanks{ J. Hribar is with CONNECT, Trinity College Dublin, Ireland. Email: jhribar@tcd.ie.

A. Marinescu is with Eaton's Center for Intelligent Power, Dublin, Ireland. Email: AndreiMarinescu@eaton.com.

A. Chiumento is with the University of Twente, The Netherlands, Email: a.chiumento@utwente.nl.

and L. A. DaSilva is with the Commonwealth Cyber Initiative, Virginia Tech, USA. Email: ldasilva@vt.edu.

This work was funded in part by the European Regional Development Fund through the SFI Research Centres Programme under Grant No. 13/RC/2077\_2 SFI CONNECT and by the SFI-NSFC Partnership Programme Grant Number 17/NSFC/5224. 
The corresponding author is Jernej Hribar.}
}

\maketitle
\copyrightnotice


\begin{acronym}[MACHU]
  \acro{iot}[IoT]{Internet of Things}
  \acro{cr}[CR]{Cognitive Radio}
  \acro{ofdm}[OFDM]{orthogonal frequency-division multiplexing}
  \acro{ofdma}[OFDMA]{orthogonal frequency-division multiple access}
  \acro{scfdma}[SC-FDMA]{single carrier frequency division multiple access}
  \acro{rbi}[RBI]{ Research Brazil Ireland}
  \acro{rfic}[RFIC]{radio frequency integrated circuit}
  \acro{sdr}[SDR]{Software Defined Radio}
  \acro{sdn}[SDN]{Software Defined Networking}
  \acro{su}[SU]{Secondary User}
  \acro{ra}[RA]{Resource Allocation}
  \acro{qos}[QoS]{quality of service}
  \acro{usrp}[USRP]{Universal Software Radio Peripheral}
  \acro{mno}[MNO]{Mobile Network Operator}
  \acro{mnos}[MNOs]{Mobile Network Operators}
  \acro{gsm}[GSM]{Global System for Mobile communications}
  \acro{tdma}[TDMA]{Time-Division Multiple Access}
  \acro{fdma}[FDMA]{Frequency-Division Multiple Access}
  \acro{gprs}[GPRS]{General Packet Radio Service}
  \acro{msc}[MSC]{Mobile Switching Centre}
  \acro{bsc}[BSC]{Base Station Controller}
  \acro{umts}[UMTS]{universal mobile telecommunications system}
  \acro{Wcdma}[WCDMA]{Wide-band code division multiple access}
  \acro{wcdma}[WCDMA]{wide-band code division multiple access}
  \acro{cdma}[CDMA]{code division multiple access}
  \acro{lte}[LTE]{Long Term Evolution}
  \acro{papr}[PAPR]{peak-to-average power rating}
  \acro{hn}[HetNet]{heterogeneous networks}
  \acro{phy}[PHY]{physical layer}
  \acro{mac}[MAC]{medium access control}
  \acro{amc}[AMC]{adaptive modulation and coding}
  \acro{mimo}[MIMO]{multiple input multiple output}
  \acro{rats}[RATs]{radio access technologies}
  \acro{vni}[VNI]{visual networking index}
  \acro{rbs}[RB]{resource blocks}
  \acro{rb}[RB]{resource block}
  \acro{ue}[UE]{user equipment}
  \acro{cqi}[CQI]{Channel Quality Indicator}
  \acro{hd}[HD]{half-duplex}
  \acro{fd}[FD]{full-duplex}
  \acro{sic}[SIC]{self-interference cancellation}
  \acro{si}[SI]{self-interference}
  \acro{bs}[BS]{base station}
  \acro{fbmc}[FBMC]{Filter Bank Multi-Carrier}
  \acro{ufmc}[UFMC]{Universal Filtered Multi-Carrier}
  \acro{scm}[SCM]{Single Carrier Modulation}
  \acro{isi}[ISI]{inter-symbol interference}
  \acro{ftn}[FTN]{Faster-Than-Nyquist}
  \acro{m2m}[M2M]{machine-to-machine}
  \acro{mtc}[MTC]{machine type communication}
  \acro{mmw}[mmWave]{millimeter wave}
  \acro{bf}[BF]{beamforming}
  \acro{los}[LOS]{line-of-sight}
  \acro{nlos}[NLOS]{non line-of-sight}
  \acro{capex}[CAPEX]{capital expenditure}
  \acro{opex}[OPEX]{operational expenditure}
  \acro{ict}[ICT]{information and communications technology}
  \acro{sp}[SP]{service providers}
  \acro{inp}[InP]{infrastructure providers}
  \acro{mvnp}[MVNP]{mobile virtual network provider}
  \acro{mvno}[MVNO]{mobile virtual network operator}
  \acro{nfv}[NFV]{network function virtualization}
  \acro{vnfs}[VNF]{virtual network functions}
  \acro{cran}[C-RAN]{Cloud Radio Access Network}
  \acro{bbu}[BBU]{baseband unit}
  \acro{bbus}[BBU]{baseband units}
  \acro{rrh}[RRH]{remote radio head}
  \acro{rrhs}[RRH]{Remote radio heads} 
  \acro{sfv}[SFV]{sensor function virtualization}
  \acro{wsn}[WSN]{wireless sensor networks} 
  \acro{bio}[BIO]{Bristol is open}
  \acro{vitro}[VITRO]{Virtualized dIstributed plaTfoRms of smart Objects}
  \acro{os}[OS]{operating system}
  \acro{www}[WWW]{world wide web}
  \acro{iotvn}[IoT-VN]{IoT virtual network}
  \acro{mems}[MEMS]{micro electro mechanical system}
  \acro{mec}[MEC]{Mobile edge computing}
  \acro{coap}[CoAP]{Constrained Application Protocol}
  \acro{vsn}[VSN]{Virtual sensor network}
  \acro{rest}[REST]{REpresentational State Transfer}
  \acro{aoi}[AoI]{Age of Information}
  \acro{lora}[LoRa\texttrademark]{Long Range}
  \acro{iot}[IoT]{Internet of Things}
  \acro{snr}[SNR]{Signal-to-Noise Ratio}
  \acro{cps}[CPS]{Cyber-Physical System}
  \acro{uav}[UAV]{Unmanned Aerial Vehicle}
  \acro{rfid}[RFID]{Radio-frequency identification}
  \acro{lpwan}[LPWAN]{Low-Power Wide-Area Network}
  \acro{lgfs}[LGFS]{Last Generated First Served}
  \acro{wsn}[WSN]{wireless sensor network} 
  \acro{lmmse}[LMMSE]{Linear Minimum Mean Square Error}
  \acro{rl}[RL]{Reinforcement Learning}
  \acro{nb-iot}[NB-IoT]{Narrowband IoT}
  \acro{lorawan}[LoRaWAN]{Long Range Wide Area Network}
  \acro{mdp}[MDP]{Markov Decision Process}
  \acro{ann}[ANN]{Artificial Neural Network}
  \acro{dqn}[DQN]{Deep Q-Network}
  \acro{mse}[MSE]{Mean Square Error}
  \acro{ml}[ML]{Machine Learning}
  \acro{cpu}[CPU]{Central Processing Unit}
  \acro{ddpg}[DDPG]{Deep Deterministic Policy Gradient}
  \acro{ai}[AI]{Artificial Intelligence}
  \acro{gp}[GP]{Gaussian Processes}
  \acro{rmse}[RMSE]{Root Mean Square Error}
  \acro{drl}[DRL]{Deep Reinforcement Learning}
  \acro{mmse}[MMSE]{Minimum Mean Square Error}
  \acro{fnn}[FNN]{Feedforward Neural Network}
\end{acronym}

\begin{abstract}

Millions of battery-powered sensors deployed for monitoring purposes in a multitude of scenarios, e.g., agriculture, smart cities, industry, etc., require energy-efficient solutions to prolong their lifetime. When these sensors observe a phenomenon distributed in space and evolving in time, it is expected that collected observations will be correlated in time and space. 
This paper proposes a \ac{drl} based scheduling mechanism capable of taking advantage of correlated information. The designed solution employs \ac{ddpg} algorithm. The proposed mechanism can determine the frequency with which sensors should transmit their updates, to ensure accurate collection of observations, while simultaneously considering the energy available. The solution is evaluated with multiple datasets containing environmental observations obtained in multiple real deployments. The real observations are leveraged to model the environment with which the mechanism interacts as realistically as possible. The proposed solution can significantly extend the sensors' lifetime and is compared to an idealized, all-knowing scheduler to demonstrate that its performance is near-optimal.
Additionally, the results highlight the unique feature of proposed design, energy-awareness, by displaying the impact of sensors' energy levels on the frequency of updates.

\end{abstract}

\acresetall


\section{Introduction}
\label{sec:intro}

Millions of low-power devices are being deployed to provide services in smart cities \cite{zanella2014internet}, Industry 4.0 \cite{da2014internet}, smart agriculture \cite{wolfert2017big}, and other \ac{iot} applications. Many of these devices are low-cost sensors powered by non-rechargeable batteries. Their role is to provide sensed information to services, which use this information to make decisions. For example, in smart agriculture, a service controlling an irrigation system requires information from various sensors to decide which fields to water. The main challenge is to provide accurate and up-to-date information to services while keeping the deployments of battery-powered devices functional for as long as possible.

In a system of multiple sensing devices observing the same physical phenomenon, it is expected that the information collected will be correlated in time and space. By relying on this correlation, we have shown that it is possible to increase the time between consecutive updates by each sensor, thereby increasing its lifetime, without compromising the accuracy of the information provided to the \ac{iot} service~\cite{hribar2018using}. In the absence of up-to-date information from one sensor, the system can rely on more recent information obtained from a correlated sensor. In this paper, we propose a \ac{drl}-based scheduling mechanism capable of determining how frequently each low-power sensor should transmit its observations so as to furnish the service with accurate information while maximising the lifetime of the network.

We consider an \ac{iot} system where low-power sensors transmit periodic updates to a gateway. The gateway is able to schedule when the next update by each sensor should occur. The gateway relies on a data-driven approach to make this determination, by considering the energy available to each low-power sensor and the need for fresh updates, according to concepts related to the \ac{aoi}\cite{kosta2017age}. We design a \ac{drl} solution capable of learning from past experiences.

Multiple, often non-trivially connected, factors impact the decision of when a particular low-power sensor should transmit new information. These factors can be external to the sensor, such as whether a nearby sensor has recently transmitted updated information, changes in the observed physical phenomenon, etc., or internal to the sensor, e.g., the remaining energy, transmission power, or location. The use of \ac{drl} enables us to design a scheduling mechanism capable of determining when sensors should transmit updated sensed information to a gateway, by weighing all relevant factors to make an efficient decision.

In this paper, we use \ac{drl} to conserve battery-powered sensors' energy by leveraging the correlation exhibited in the information they collect. In particular, we make the following contributions: 
\begin{itemize}
\item The main contribution of this paper is the design of a \ac{drl} energy-aware scheduler that is capable of determining when an \ac{iot} sensor should transmit its next observation. We make use of \ac{ddpg}, a \ac{drl} algorithm, to arrive at an efficient transmission schedule for sensors based on their available energy, the freshness of the information collected, and the expected lifetime of all other sensors in the network, without compromising the accuracy of the information delivered to the application. 
\item A unique feature of our solution is energy balancing. Our mechanism is capable of determining to what extent the energy available to one sensor can be used to prolong the lifetime of others. We benchmark our solution by comparing it to an ideal scheduler that acts as an oracle and is assumed to know the ground truth about the phenomenon being observed.
\item To validate our solution, we use more than five different datasets to demonstrate its near-optimal performance in a variety of scenarios. Note that we leverage real data to model the environment in which our scheduler operates as realistically as possible.
\end{itemize}

After briefly reviewing the relevant literature in Section~\ref{sec:related}, in Section~\ref{sec:system_model} we describe how  a system of sensors collecting correlated information can estimate the accuracy of their observations. We describe the decision-making problem that our proposed scheduling mechanism is capable of assisting with in Section~\ref{sec:problem_formulation}. We present how we incorporated \ac{ddpg} into our proposed mechanism (Section \ref{sec:rl_approach} A), describe the system dynamics using states, actions, and rewards from a \ac{rl} perspective (Section~\ref{sec:rl_approach} B), and provide implementation details (Section~\ref{sec:rl_approach} C). We utilize data obtained from real deployments to show that the learned behaviour significantly prolongs the sensors' lifetime and achieves near-optimal performance (Section~\ref{sec:validation} A). Additionally, we demonstrate the scheduling mechanism's energy awareness when deciding on the sensors' transmission times (Section \ref{sec:validation} B). Finally, we summarise our main findings and discuss our future work in Section \ref{sec:conclusion}.

\section{Related Work}
\label{sec:related}
Our work leverages the recently proposed \ac{aoi} metric~\cite{kaul2011piggybacking}, which quantifies the freshness of information. The \ac{aoi} metric measures the time since a status update generated by a source, e.g., a sensor, was last received. Each status update contains the latest information collected by the source and a timestamp. The more recently the status update was generated, the more relevant it is to the decision process. Finding the optimal update rates with which sources should send information is non-trivial~\cite{yates2015lazy}, and considering correlation between status updates adds to the complexity of the problem, as an update from one source lowers the requirement for fresh information on all other correlated sources. We were the first to propose taking advantage of fresher information from correlated sources to conserve the energy of battery-powered sources by prolonging the times between sources' consecutive updates~\cite{hribar2018using}. Subsequently, the authors in \cite{kalor2019minimizing} and \cite{jiang2019status} analyzed from different perspectives a system with correlated sources. In \cite{kalor2019minimizing}, the authors consider a system in which multiple sources may observe the same information captured by sensors, i.e., multiple sources can obtain the same status update. In \cite{jiang2019status}, the authors considered a one-dimensional static random field from which multiple sources transmitted observations. The works in \cite{hribar2018using, kalor2019minimizing, jiang2019status}  considered only the impact of correlation on the timeliness of information to establish the desired frequency of updates. This paper presents a solution that considers both the timeliness of the information and the energy available to the sources to arrive at a scheduling of information updates that prolongs the lifetime of battery-powered sensors. 

Our work is also related to, but differs in crucial aspects from, the various approaches proposed for energy efficiency in the context of \acp{wsn} \cite{anastasi2009energy, rault2014energy, villas2014spatial, yetgin2017survey, carrano2014survey}. Most proposed works for \acp{wsn} rely on detection or reconstruction of the observed phenomena, through data prediction or model-based active sampling methods, to improve the low-power sensors' energy efficiency. In contrast, we focus on the timeliness of updates, and in particular the \ac{aoi}, and then employ \ac{rl} to determine how to utilize correlated measurements to reduce the rate at which sources transmit their updates.

\ac{rl} has been applied to other energy-aware networking solutions~\cite{luong2018applications, li2018q, mohammadi2018semisupervised, alsheikh2014machine, zheng2015green, aoudia2018rlman, du2018deep, zhu2018new, ning2019deep, sharma2019distributed}. 
For example, in \cite{li2018q} the authors use Q-learning to enhance the spectrum utilization of industrial \ac{iot} devices. They demonstrate that devices are capable of learning a channel selection policy to avoid collisions, thus reducing retransmissions. The authors in \cite{zhu2018new} propose a channel-state aware scheduling mechanism using a \ac{dqn} that enables a device to learn when to transmit and which channel to use for transmission. The authors in \cite{mohammadi2018semisupervised} investigate the use of a semi-supervised \ac{dqn} to improve indoor users' location estimation by leveraging information on Bluetooth signal strength. The work in~\cite{zheng2015green} relies on a no regret \ac{rl} approach, while the authors in~\cite{aoudia2018rlman} apply an actor-critic algorithm to analyse how an energy-harvesting device will collect energy and schedule its transmissions accordingly. In both cases, the objective is more effective power management, using \ac{rl} to prevent power outages, i.e., to avoid the situation where an energy-harvesting device completely depletes its energy. The power control in an energy harvesting system is also investigated in \cite{sharma2019distributed}. The authors, using a multi-agent approach (with a \ac{dqn} algorithm), designed a distributed power control mechanism that maximises devices' throughput. The authors in \cite{ning2019deep} use a \ac{dqn} algorithm to solve a task offloading problem for vehicles. Their solution is capable of saving energy by selecting more efficiently where in the edge a task from a vehicle should be processed. In \cite{du2018deep}, the authors apply a deep belief neural network to design a data recovery mechanism for sensors that collect spatio-temporally correlated information. Their mechanism is capable of determining which observations from other sensors could be used to replace missing or corrupted observations. 

In \cite{hribar2019ICC}, we first introduced a variant of the proposed scheduling mechanism. In this paper, we extend our work by designing an improved \ac{drl}-based mechanism (based on a \ac{ddpg} algorithm, while~\cite{hribar2019ICC} relied on a \ac{dqn} algorithm) and providing a comprehensive evaluation of the effectiveness of our solution. We compare our solution to one conventional way of setting the sensors' update intervals, and to the optimal scheduler we designed specifically for evaluation purposes. Furthermore, we test our mechanism over three additional real-world datasets. Another extension of our work is in providing new results that demonstrate the energy balancing aspect of our solution.


\section{Quantifying the accuracy of observations}
\label{sec:system_model}

We consider a sensor network with $N$ geographically distributed sensors transmitting observations to a gateway for collection. The main purpose of these sensors, denoted as $\{S_{1}, \hdots, S_{N}\}$, is to observe a physical phenomenon $Z(\bm{X},t)$ distributed in space $\bm{X}$ and evolving in time $t$. In our work, we perform the physical modeling of the observed phenomenon using observations obtained in a real \ac{iot} deployment \cite{bodik2004intel, gutierrez2016co}. Sensors are deployed at positions $\bm{X}_{n}$ and transmit periodic observations with an update interval $T_n, n=1, \ldots, N$. In our system, we assume that the latest received observation from a sensor replaces the previously received information as, according to the \ac{aoi} paradigm, the freshest information is the most relevant in the decision making process \cite{kaul2011piggybacking}. The \ac{aoi} metric measures the time elapsed since the sink, i.e., a gateway, received a new observation from the sensor. We denote the \ac{aoi} with $\Delta_{n}(t)$. Whenever the system receives an observation from location $\bm{X}_{n}$ at time $t_{n}$, the system will anticipate the arrival of the next observation from location $\bm{X}_{n}$ at time instance $t = t_n + T_n$. Additionally, the value of \ac{aoi} will drop to zero. Meaning, that the value of \ac{aoi} is limited to an interval between $0$ and $T_n$. We write the collected observations into a vector $\mathbf{Y} = \left[y_1, \ldots, y_{N} \right]^T$ with $y_n = Z \left(\bm{X}_{n}, t_{n} \right)$ where $t_n$ is the latest time at which sensor $n$ has reported an observation. 

\begin{figure} 
	\centering
	\resizebox {0.65\columnwidth} {!} {
\begin{tikzpicture}
	\begin{pgfonlayer}{main}
	\draw[line width=0.5mm](-2,3) circle (2pt);
	\node  at (-1.75,3.25) {$S_n$};
	\draw[thick,<->,line width=0.5mm] (-1.95,2.90) -- (-1.02,0.1);
	\node  at (-1.3,1.55) {$d_{n,i}$};
	\draw[line width=0.5mm] (0,0) circle (5pt);
	\node[cross, line width=0.6mm,minimum size=5pt] at (0,0) {};
	\node  at (-1,-0.35) {$Z(\bm{X}_i,t)$};
	\draw[fill=white] (0.7,1.75) -- (0.7,3.25) -- (3.5,3.25) -- (3.5,1.75) -- (0.7,1.75);
	\draw[line width=0.5mm,fill=white](1,3) circle (2pt);
	\node  at (2.35,3) {\footnotesize IoT Sensor};
	\draw[line width=0.5mm,fill=white](1,2.5) circle (5pt);
	\node[cross, line width=0.6mm,minimum size=5pt] at (1,2.5) {};
	\node  at (2.35,2.5) {\footnotesize Gateway};
	\draw[gray!100, fill] (1,2) circle (2pt);
	\node  at (2.35,2) {\footnotesize Estimation Point};
	
	\draw[line width=0.5mm](-1.5,-1.6) circle (2pt);
	\node  at (-1.1,-1.6) {$S_1$};
	\draw[line width=0.5mm](-3.5,0.3) circle (2pt);
	\node  at (-3.5,-0.1) {$S_N$};
	\draw[line width=0.5mm](-1,0) circle (2pt);
	
	\draw[line width=0.5mm](-1.7,1.1) circle (2pt);
	\draw[line width=0.5mm](-2.25,-0.6) circle (2pt);
	\draw[line width=0.5mm](-2.85,-1.9) circle (2pt);
	\draw[line width=0.5mm](2.6,-0.7) circle (2pt);
	\draw[line width=0.5mm](-2.2,2.6) circle (2pt);
	
	\draw[line width=0.5mm](1.9,0.6) circle (2pt);
	\draw[line width=0.5mm](-3,2) circle (2pt);
	\draw[line width=0.5mm](1.5,-1) circle (2pt);
	\draw[line width=0.5mm](0.1,2.8) circle (2pt);
	\draw[line width=0.5mm](-3.5,-2.2) circle (2pt);
	
	\draw[line width=0.5mm](1.1,-0.7) circle (2pt);
	\draw[line width=0.5mm](2.1,-1.7) circle (2pt);
	\draw[line width=0.5mm](-2.85,1.2) circle (2pt);
	\draw[line width=0.5mm](-1.82,-2.6) circle (2pt);
	\draw[line width=0.5mm](-0.3,2,8) circle (2pt);
	
	\draw[line width=0.5mm](0.1,-2.8) circle (2pt);
	\draw[line width=0.5mm](-0.6,2.1) circle (2pt);
	\draw[line width=0.5mm](0.3,0.8) circle (2pt);
	\end{pgfonlayer}
	
	\begin{pgfonlayer}{background}
	circles 
	\draw[gray!100, fill] (-1,0) circle (2pt);
	\draw[dashed,gray!95] (-1,0) circle (10pt);
	\draw[dashed,gray!85] (-1,0) circle (25pt);
	\draw[dashed,gray!70] (-1,0) circle (40pt);
	\draw[dashed,gray!55] (-1,0) circle (55pt);
	\draw[dashed,gray!40] (-1,0) circle (70pt);
	\draw[dashed,gray!25] (-1,0) circle (85pt);
	\draw[dashed,gray!20] (-1,0) circle (100pt);
	\end{pgfonlayer}
\end{tikzpicture}
	}
	\caption{A system of $N$ randomly distributed sensor nodes, whose observations are used to estimate the value of observed physical phenomenon, $Z(\bm{X},t)$, at location $\bm{X}_i$  at time $t$.}
	\label{fig:system}
	\vspace{-10pt}
\end{figure}
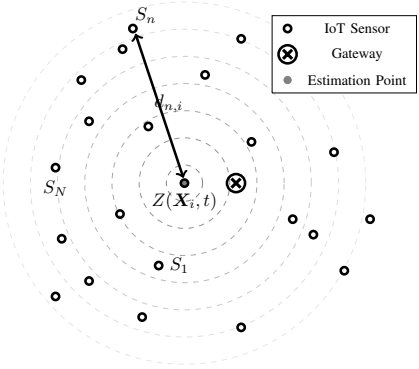

The system can estimate the value of the observed physical phenomenon at the desired location $\bm{X}_i$ at any time instant $t$ using the collected information, as presented in Fig \ref{fig:system}. We denote the Euclidean distance between sensor $S_n$ and the location of interest $\bm{X}_i$ as $d_{n,i}$. With $\Delta_{n,i}(t)$  we denote the time elapsed since the system received the latest observation from sensor $S_n$, i.e., the \ac{aoi}, $\Delta_n(t)\coloneqq t-t_n$. To estimate the observed process we apply a \ac{lmmse} estimator which is commonly used for such problems as demonstrated in \cite{schizas2008consensus}. One of the biggest advantage of using \ac{lmmse} is that it provides a mathematically tractable solution and to obtain it, we only require the expected values, variances, and the covariance. In our system, all three are available. Therefore, we can approximate the value of the observed physical phenomenon at position $\bm{X}_i$ at time instant $t$, as:

\begin{equation}\label{eq:estimation}
\hat{y}_i(t) = \sum_{n=1}^{N} w_{n, i}(t) y_n,
\end{equation}

\noindent where $w_n(t), n=0, \ldots , N$ are \ac{lmmse} estimator weights.

Following the analysis in~\cite{oppenheim2015signals}, we obtain the \ac{lmmse} estimator weight vector $\boldsymbol{W}_i(t) = \left[w_{0,i}(t), \ldots, w_{N, i}(t) \right]^T$ as follows:

\begin{equation}\label{eq:lmmse}
\boldsymbol{W}_i(t) = \bigg (\bm{C}_{\scriptstyle \bm{YY}} (t) \bigg)^{-1} \bm{c}_{\scriptstyle i, \bm{Y}Z} (t) \displaystyle.
\end{equation}

\noindent The matrices $\bm{C}_{\scriptstyle \bm{YY}} (t), \bm{c}_{\scriptstyle i, \bm{Y}Z}(t) $ are covariance matrices, required to determine $\boldsymbol{W}_i(t)$:

\begin{equation}\label{eq:covariance_matrix}
\bm{C}_{\scriptstyle \bm{YY} }(t) =\begin{bmatrix}
\mathrm{C}_{1,1}(t) \hdots  \mathrm{C}_{1,N}(t)   \\
\vdots \phantom{ABCDEF}  \vdots \\
\mathrm{C}_{N,1}(t)  \hdots \mathrm{C}_{N,N}(t) \\
\end{bmatrix};
\bm{c}_{\scriptstyle i, \bm{Y}Z}(t) =
\begin{bmatrix}
\mathrm{C}_{1,i}(t) \\
\vdots \\
\mathrm{C}_{N,i}(t) \\
\end{bmatrix};
\end{equation}

\noindent in which $\mathrm{C}_{j,k}(t) ; j,k=1, \ldots, N $, is the covariance of observations $y_j$ and $y_k$, and $\mathrm{C}_{j,i}(t)$ is the covariance of $y_j$ and the observed process $Z$ at the desired location of the estimation. To obtain the required matrices we can rely on a covariance model and utilize past observations to determine its values. We adopt a separable covariance model defined in~\cite{cressie1999classes}. With it we model how observations collected at different instants in time and different locations relate to each other. We express the covariance between two observations or one observation and the estimation point, with \ac{aoi} difference $\Delta_{j,k}(t)$ and distance $d_{j,k}{}$ apart as:

\begin{equation}\label{eq:correlation_II}
\mathrm{C}_{j,k}\big (d_{j,k},t| \theta_1(t), \theta_2(t) \big ) =   exp(-\theta_2(t) d_{j,k} -\theta_1(t) \Delta_{j,k}(t)).
\end{equation}

\noindent Note that $\theta_1(t)$ and $\theta_2(t)$ are scaling parameters of time and space, respectively. With $d_{j,k}$ we denote the Euclidean distance between sensor $S_j$ and the location at which the system estimates the value of the observed physical phenomenon, or between sensor $S_j$ and sensor $S_k$. Both scaling parameters change over time and are extracted from the obtained observations. In our work, we follow a scaling extraction method with Pearson's correlation coefficient formula for samples, as described in \cite{gneiting2002nonseparable}.

The selected covariance function provides a good fit to model spatial and temporal variation for many physical phenomena. For example, in~\cite{cressie1999classes}, the authors showed that such a covariance model could be applied to wind-speed data. We demonstrated in~\cite{hribar2018using} that the selected model is applicable to the temperature and humidity sensor data used in the evaluation section. Additionally, such spatio-temporal correlation can be observed in many \ac{iot} sensor deployments: examples include \ac{iot} systems in a smart city measuring air pollution, precipitation, or noise~\cite{zanella2014internet}, and smart farm applications in which an \ac{iot} system monitors soil parameters~\cite{wolfert2017big}. 

Every time the system employs Eq. \eqref{eq:estimation} to estimate the value of the observed physical phenomenon it makes an error. By using matrices $\bm{C_{\scriptstyle YY}} (t)$ and $\bm{c}_{\scriptstyle \bm{Y}Z}(t) $ it is possible to determine the \ac{mse} in the estimation as:

\begin{equation}\label{eq:system_estimatio_error_n}
\varepsilon_i \big (\bm{X}_i,t| \theta_1(t), \theta_2(t) \big ) = \sigma^2_{Z} - \bm{c}_{\scriptstyle Z\bm{Y}}(t) \boldsymbol{W}_i (t),
\end{equation}

\noindent where $ \bm{c}_{\scriptstyle Z\bm{Y}}$ is the transpose of $\bm{c}_{\scriptstyle \bm{Y}Z} $ defined above, and $\sigma^2_{Z}$ represents the variance of the observed phenomenon. The estimation error provides a measure with which the gateway can quantify the quality of the information currently provided by the sensing process: the lower the value of the estimation error, the more accurate the estimated values and the lower the need for an additional update. Hence, by measuring the average estimation error between two consecutive updates the gateway can assess how accurate and up-to-date the observations collected by the system are. 
In our work, we control the accuracy of our sensing process by setting as a constraint the maximum \ac{mse} of the estimator, $\varepsilon^*$. In short, the purpose of our proposed updating mechanism is to set sensors' update intervals in such a way that the average estimation error will not exceed the set target. In the next section, we describe the optimisation problem that the gateway must solve: maximising the network lifetime, constrained by the target accuracy in the measurements.

\section{Problem Formulation}
\label{sec:problem_formulation}

The scenario of interest to our work is the use of inexpensive battery-powered sensors, transmitting observations to a gateway for collection. The gateway aims to schedule the transmission of observations in such a way that the accuracy of the information collected will satisfy the service requirements, i.e., average estimation error below $\varepsilon^*$, while, simultaneously, trying to prolong the lifetime of the deployed sensor network. Whenever a sensor transmits an observation, the updating mechanism residing in the gateway decides on the low-power sensors' next update time by evaluating the accuracy of collected observations and the sensors' available energy. Each sensor's lifetime depends on the frequency of transmitted observations, i.e., the time between two consecutive updates, and on the continuous power consumption, which is independent of transmissions.

In this work, we assume that a non-rechargeable primary battery powers the low-power sensors. In such a case, the energy consumption depends on how often a sensor transmits an observation and the energy it needs to function regardless of the mode of operation.
Therefore, we can model a sensor's lifetime $\mathcal{L}_n(T_n) $ as in~\cite{chen2005lifetime}:

\begin{equation}\label{eq:lifetime}
\mathcal{L}_n(T_n) = \frac{{E}_0}{P_c + \frac{E_{tr}}{T_n}},
\end{equation}

\noindent where ${E}_0$ represents the sensors' starting energy and $P_c$ is the continuous power consumption, and $E_{tr}$ represents the energy required to acquire and transmit the observation. The continuous power consumption is the power that the sensor always consumes, for example, leakage current, and depends solely on the sensor hardware components. For low-power \ac{iot} sensors,  $P_c$ is in the range of a few $ \mu W$ or less. The energy required to transmit the observation, i.e., $E_{tr}$, depends on many factors such as the size of the transmitted packet, the energy required to take the measurement, and channel conditions. 

\begin{figure}
	\centering
	\includegraphics[width=3.5in]{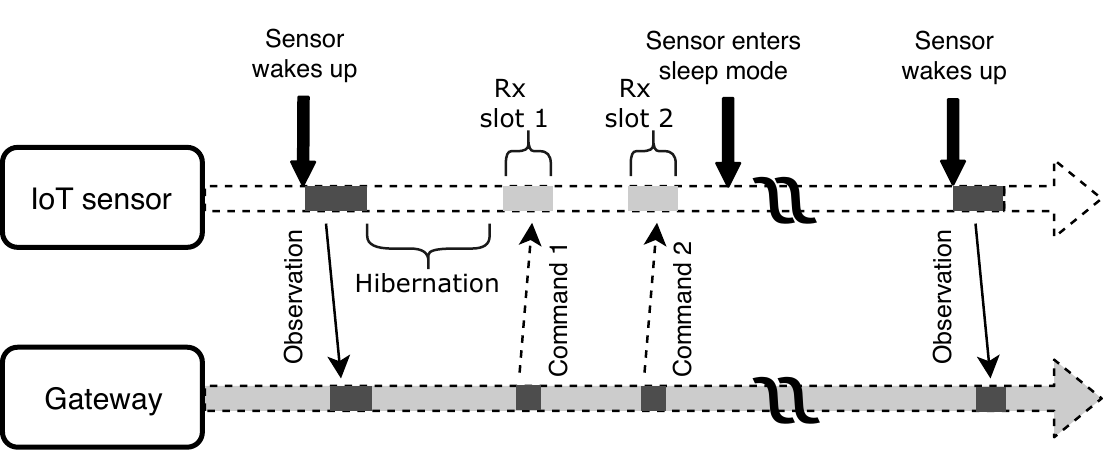}
	\caption{Message sequence of a low-power sensor using \ac{lorawan}.}
	\label{fig:Lora_message_sequence}
\end{figure}

Energy is not the only factor the updating mechanism has to take into account. As described in \cite{bormann2014terminology}, low-power sensors are also constrained in terms of available computing power, memory, communication capabilities, etc. Limited processing power and memory prevent the use of a sophisticated algorithm on the sensor itself. Therefore, computationally demanding tasks when making a decision should be carried out at the gateway. Additionally, these sensors rely on low data rate transmission, meaning that communication messages between sensors and gateway should be kept to a minimum. Furthermore, to extend their lifetime, low-power sensors rely on the use of sleep mode. When a sensor is in sleep mode, the rest of the network cannot communicate with it. Consequently, the gateway has to inform each sensor, while the sensor is still in active mode, when it should wake up again and transmit the next observation. Sleep mode is supported by most \ac{lpwan} standards, such as SigFox, Weightless, \ac{lorawan}\footnote{For more information, the reader may visit http://www.sigfox.com; http://weightless.org; http://www.lora-aliance.org, respectively.}, and \ac{nb-iot}\cite{tsoukaneri2018group}. The low-power sensor is usually in active mode only after it has transmitted. For example, a sensor using a \ac{lorawan} class A radio will listen for two short time-windows after it has transmitted, as illustrated in the \ac{lorawan} message sequence in Fig. \ref{fig:Lora_message_sequence} \cite{loradatasheet}, meaning that the updating mechanism only has a short time-window to provide a response. 

\begin{figure}
	\centering
	\includegraphics[width=3.3in]{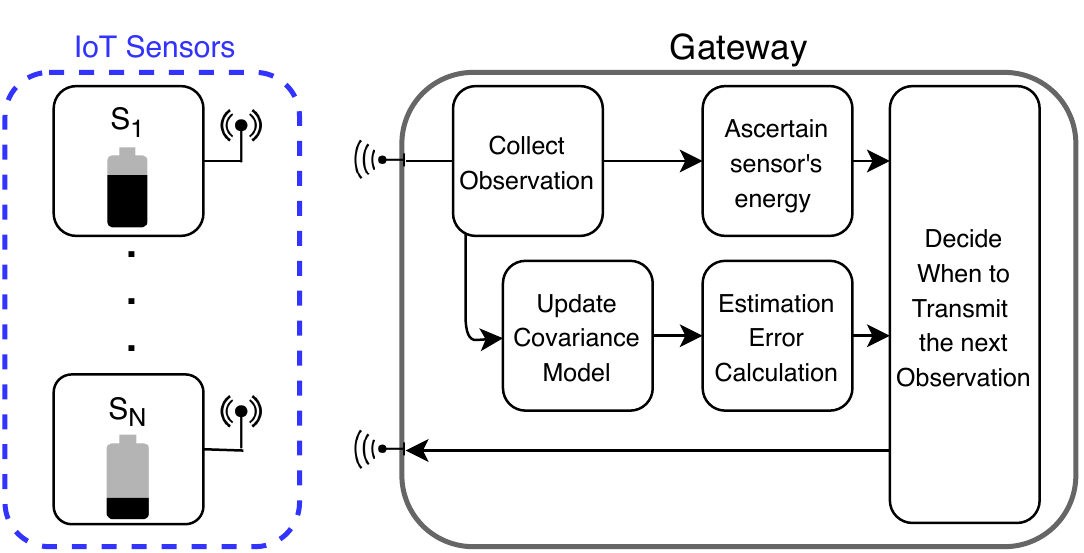}
	\caption{High level overview of the decision making process in the gateway.}
	\label{fig:problem_overview}
	\vspace{-10pt}
\end{figure}

The gateway's goal is to prolong the network lifetime. We define the network lifetime as the lifetime of the sensor with the shortest lifespan in the deployment. In other words, the network lifetime expires the moment one sensor depletes all of its energy. To that end, the gateway should aim to minimise transmissions by all sensors, i.e., increase $T_n$, and, when updates are required, favour sensors with higher remaining energy, all while keeping the average estimation error of the observed physical phenomenon at every location of interest below the pre-specified value, i.e., $\varepsilon^*$. In a real deployment, services dictate which locations are of interest for the system. In this paper, we consider every sensor location, i.e., $\bm{X}_{n}$, to be a location of interest, meaning that system has to make accurate estimations at the location of every sensor while keeping sensors' power consumption to a minimum. We summarise the decision-making process in Fig. \ref{fig:problem_overview}. The gateway decides on each sensor's next update time by evaluating the accuracy of the collected observation and the sensors' available energy, which it can determine from the sensor's reported power supply measurement. The gateway can then decide when the sensor should transmit its next observation.

Transmission from a sensor $\hat{n}$ starts the decision-making process at time $t_{\hat{n}}$: due to the energy-constrained nature of the devices, the gateway has a limited time to determine the sensor's next update time, i.e., determine for sensor $\hat{n}$ a new sensor update interval $T_{\hat{n}}$. In practice, the gateway usually has around one second to reply with the new update interval. Further, we discretise the possible update intervals in steps of duration $t_S$, up to a maximum update interval of $T_{max}$, an integer multiple of $t_S$. Discretising time enables us to formulate the problem the gateway faces as follows:

\begin{maxi}|l|
    {T_{\hat{n}}}{
    \min  \bigg( \mathcal{L}_n (T_n)\bigg),  \forall \: n=1,\ldots,N  }{}{}
    \label{eq:problem_formulation}
    \addConstraint{\varepsilon_n(t_{\hat{n}} + T_{\hat{n}}) }{\leq \varepsilon^*, \forall \: n=1,\ldots,N.}{}
    \addConstraint {T_{\hat{n}} \in \{t_S, 2t_S, \ldots, T_{max}\}}{}.
\end{maxi}
\noindent 

At time step $t_{\hat{n}}$, when the gateway receives an observation from sensor $\hat{n}$, the gateway has to select the update interval $T_{\hat{n}}$ that will maximise the lifetime of the sensor (or sensors) in the deployment with the shortest life expectancy. The gateway has to ensure that the accuracy constraint, i.e., $\varepsilon_n \leq \varepsilon^*$, is met for all sensors until the next update is received at time $t_{\hat{n}} + T_{\hat{n}}$. In other words, until the system receives a new update from sensor $\hat{n}$, the value of the estimation error should not exceed the set target on any of the sensors. Additionally, because one sensor's update time potentially affects the lifetime of all other sensors, the gateway might have to select a lower update time for sensor $\hat{n}$. By doing so, the scheduler can prolong the lifetime of another sensor, preferably one with less available energy, to maximise the lifetime, i.e., $\mathcal{L}_n (T_n),  \forall \: n=1,\ldots, N$.

To solve the problem, the gateway could turn to a numerical solver. However, due to the matrix inversion in Eq. \eqref{eq:lmmse}, required to determine the \ac{mse} value using Eq. \eqref{eq:system_estimatio_error_n} for $N$ sensors, the expected computational complexity to solve the proposed problem is of the order of $\mathcal{O}(N^4 \frac{T_{max}}{t_S})$. The required computational power for a conventional numerical solution rises to the power of four with the number of sensors. In practice, the gateway does not have enough time available to determine the sensor's new update interval. Additionally, if conditions in the environment change, e.g., a sensor disconnects, the optimization problem becomes absolute. Everything considered the most practical approach is to employ a \ac{drl} algorithm. The \ac{drl} removes the constraint of a computationally intense on-line method as all the power is used in off-line training. Consequently, the system can respond in a few milliseconds. The use of \ac{ann} also allows the system to create a model, free of human influence, for the complex relationships between the energy and correlation. Thus the learned updating strategies are not limited by design choices. Another advantage of \ac{drl} is adaptability to environmental conditions changes due to its on-line learning approach. 

The system is highly dynamical, as each received observation impacts the covariance model's scaling parameters. As a result, the value of the \ac{mse} (Eq. \eqref{eq:system_estimatio_error_n}) continuously varies over time. Intuitively, when a sensor has more energy available than others, it should transmit more often, to enable other sensors (which will transmit in the future) to increase their update intervals. Such a problem is ideal for a \ac{rl} approach because the agent (in our case the gateway) can learn how to take actions that might bring negative reward in the near future but will ultimately increase the long-term reward~\cite{sutton1998reinforcement} in prolonging the network lifetime. 
In the next section, we show how we model the decision making discussed using an \ac{rl} solution, describing the relevant states, actions, and rewards. Then, by applying a \ac{ddpg} algorithm, the gateway can arrive at a long-term updating policy to collect accurate observations and prolong the network lifetime.


\section{Deep Reinforcement Learning Approach}
\label{sec:rl_approach}

In \ac{drl}, the agent learns its optimal behaviour, i.e., the best long term action to be taken in every state, from interactions with the environment. In our case, the learning agent resides in the gateway. The agent follows a sequence of events. Once a sensor transmits an update, the agent has to respond by setting the sensor's update interval, and by calculating the \ac{mse} value it can assess the impact of the set update interval. The agent must also consider the remaining energy available in each battery-powered sensor and their current update intervals when making the decision. To solve the decision-making problem the agent is facing, we employ a \ac{ddpg} algorithm. Using the \ac{ddpg} algorithm enables us to design a scheduling mechanism with a high number of possible actions regarding setting the sensor's update interval. There are two main advantages of using a \ac{ddpg} algorithm over other \ac{drl} algorithms: 1) there is a high convergence guarantee even when using non-linear function approximations, e.g., \ac{ann}~\cite{bhatnagar2008incremental} and 2) \ac{ddpg} is deterministic, meaning that the policy gradient is integrated only over the state space, thus requiring much fewer samples to find the optimal policy in comparison to stochastic algorithms \cite{ddpg}. In our work we follow the \ac{ddpg} algorithm implementation as presented in~\cite{lillicrap2015continuous}.

\subsection{Deep Deterministic Policy Gradient Algorithm}

\ac{ddpg} is an actor-critic algorithm and, as the name suggests, consists of two entities/\acf{ann}: the actor taking actions, and a critic that evaluates them. The critic is implemented as a \ac{dqn} and we denote its \ac{ann} as $Q(\mathbf{s}, a|\theta^Q)$ where $\theta^Q$ are weights of the critic's \ac{ann}, with $a$ denoting the action the agent takes in state $\mathbf{s}$. Next, we define the actor as a parametric function $\mu(\mathbf{s}|\theta^\mu)$ in which $\theta^\mu$ represents the actor's \ac{ann} weights. In addition, during the training process we initialize the target \ac{ann} $Q^{\prime}(\mathbf{s}, a|\theta^{Q^{\prime}}) $ and $\mu^{\prime}(\mathbf{s}|\theta^{\mu^{\prime}})$ with weights $\theta^{Q^{\prime}} $ and $\theta^{\mu^{\prime}}$ for critic and actor respectively. The agent selects actions according to its current policy with added noise $a = \mu(\mathbf{s}|\theta^\mu) + \mathcal{N} $, where  $ \mathcal{N}$ represent added random noise. Note that exploration in a \ac{ddpg} algorithm is carried out by adding a random value, i.e., noise, to the actor's selected value. Then the agent transitions into a new state $\mathbf{s}'$ and receives reward $r$. The transition $(\mathbf{s}, a, r,\mathbf{s}')$, also referred to as experience, is stored in memory. When the algorithm is training the \ac{ann} it first samples a mini-batch of $M$ experiences from the batch and calculates the target values:

\begin{equation}
h_m = r_m + \gamma Q'(\mathbf{s}_m', \mu'(\mathbf{s}_m|\theta^{\mu^{\prime}})|\theta^{Q^{\prime}}),
\end{equation}
\noindent with $m$ denoting the selected experience from the batch. After determining $h_m$ we can update the critic by minimizing the loss $L$ as:
\begin{equation}
L = \frac{1}{M} \sum_{m=1}^{M}(h_m-Q(s_{m}, a_{m}\vert\theta^{Q}))^{2}.
\end{equation}
\noindent With the loss function calculated, the algorithm then updates the actor's policy using the sampled policy gradient:
\begin{equation}
\nabla_{\theta^{\mu}}J\approx\frac{1}{M}\sum_{m=1}^{M}\nabla_{a}Q(\mathbf{s},a|\theta^{Q})\vert_{\mathbf{s}=\mathbf{s}_m,a=\mu(\mathbf{s}_m)}\nabla_{\theta^{\mu}}\mu(\mathbf{s}\vert \theta^{\mu})\vert _{\mathbf{s}_m}. 
\end{equation}
\noindent In the last step the \ac{ddpg} algorithm updates the target \acp{ann}:
\begin{equation}
\theta^{Q^{\prime}}\leftarrow\tau\theta^{Q}+(1-\tau_C)\theta^{Q^{\prime}},
\end{equation}
\begin{equation}
\theta^{\mu^{\prime}}\leftarrow\tau\theta^{\mu}+(1-\tau_A)\theta^{\mu^{\prime}},
\end{equation}
\noindent where $\tau_A$ and $\tau_C$ represent the target networks update factor. Note that the role of the target \acp{ann} is to calculate $h_m$. Using separate target networks along with the replay buffer provide stability during the training process as was established in \cite{mnih2013playing}.

\begin{figure}
	\centering
	\includegraphics[width=3.3in]{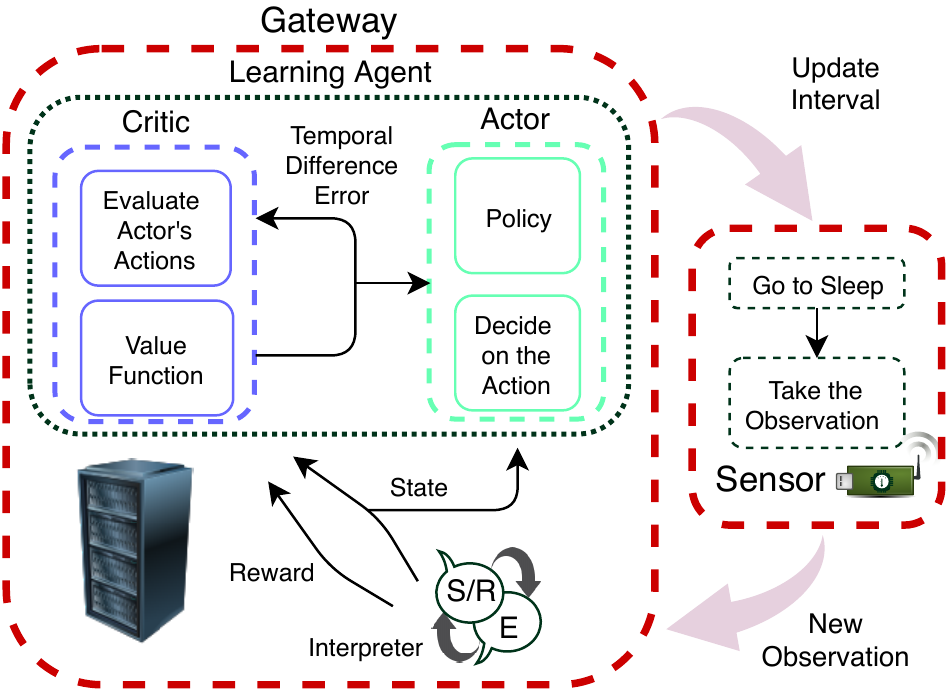}
	\caption{A high-level overview of the proposed scheduling mechanism implemented with a Deterministic Policy Gradient Algorithm.}
	\label{fig:DDPG}
\end{figure} 

Figure \ref{fig:DDPG} illustrates how we adopt \ac{ddpg} in our scheduling algorithm. In our approach, the gateway performs every computationally demanding task. A low-power sensor only receives a command message that instructs the sensor for how long it should enter sleep mode. By setting the duration of each sensor's sleep mode, the gateway effectively schedules the sensors' transmission of updates. In our mechanism the actor's actions are limited to increasing and decreasing the sleep time, and the critic's role is to evaluate whether the selected change is beneficial to the system. The critic derives a value representing the quality of the selected action using the reward for the actor. The actor then uses the provided quality of action value to adapt its policy accordingly. 

Next, we define states, actions, and rewards, i.e., a tuple in $\langle \mathcal{S}, \mathcal{A}, R \rangle$ that enables the gateway to determine each sensor's optimal update interval.

\subsection{States, Actions, and Rewards} 

Upon the $n$-th sensor's transmission the gateway constructs the state $\mathbf{s}_n \in \mathcal{S}$. The state must capture the critical aspects of the decision-making process in the gateway: sensors update interval, available energy, and the value of the estimation error. To make the decision the gateway has to weight the $n$-th sensor status (current update time, average estimation error, and energy) against the state of all other sensors. The state $\mathbf{s}_n$ can be expressed as a six-dimensional vector:

\small
\begin{equation}\label{eq:state_space}
\begin{aligned}
\mathbf{s}_n = ( T_n, E_n, \frac{\overline{\varepsilon}_n}{\varepsilon^*}, \left(\prod _{\substack{i=1 \\ i\neq n}}^{N}T_{i}\right)^{\frac {1}{N-1}}, \\
\left(\prod _{\substack{i=1 \\ i\neq n}}^{N}E_{i}\right)^{\frac {1}{N-1}}, \left(\prod _{\substack{i=1 \\ i\neq n}}^{N} \frac{\overline{\varepsilon}_{i}}{\varepsilon^*}\right)^{\frac {1}{N-1}})
\end{aligned}
\end{equation}
\normalsize

\noindent with $N$ representing the total number of sensors under the agent's control. The first three dimensions correspond to the transmitting $n$-th sensor's update interval ($T_n$), available energy($E_n$), and the ratio between \ac{mse} and target estimation error ($\overline{\varepsilon}_n / \varepsilon^*$). Relying on the ratio enables the learning agent to perform well even if the target estimation error changes, as we demonstrate in the next section. The last three dimensions of the state vector reveal the state of all other sensors in the system. Using the geometric mean enables us to reduce the number of state inputs, while simultaneously making sure the learning agent captures the most significant information about the environment. For example, the geometric mean provides information to the agent regarding whether the energy level in the majority of the sensors is low or high. Note, that even with a limited number of dimensions for the state input vector, millions of possible different states exist. The gateway uses the interpreter to construct the state vector for the sensor which observation it has just received, and for which it is making the decision.  

The learning agent's \textit{action} is limited to either increasing or decreasing the sensors' current update interval. In our implementation, the action value ($a_n \in \mathcal{A}$) returned by the \ac{ddpg} algorithm is between $-1$ and $1$, i.e., $\mathcal{A} = [-1,1]$. To determine the sensors' new update interval, we multiply the received action value by a constant, representing the selected maximum update interval change $U_{max}$. We calculate the new update interval as follows:

\begin{equation}
T_{\hat{n}} = \min \bigg( \max \big( T_n + \lfloor U_{max} a_n \rceil, 1 \big)  , T_{max }\bigg), 
\end{equation}

\noindent where $T_n$ is the sensors' previous update interval. Note that the value of $U_{max}$ can be relatively large, e.g., hundreds of seconds, and our approach will still perform well.
Additionally, the selection of $U_{max}$ dictates the number of possible actions the agent has at is disposal, which is twice the value of $U_{max}$ in time-steps. 

We form the \textit{reward} with the learning agent's goals in mind. The learning agent has to ensure that information collection is frequent enough to maintain the freshness of the information and simultaneously try to prolong the sensors' lifetime. We express the reward as: 

\begin{equation}
r_n(\overline{\varepsilon}_n, E_n) = \phi r_{acc}(\overline{\varepsilon}_n) +  (1 - \phi)  r_{en}(E_n).
\label{eq:reward}
\end{equation}

\noindent $r_{acc}(\overline{\varepsilon}_n)$ is the reward for accurate collection of observations and  $r_{en}(E_n)$ is the reward related to the energy aspect of the problem. The weight $\phi \in [0.25, 0.75]$ controls the balance between the reward for accurate collection of observations and the sensors' energy preservation. 
We restrict the range of the weight $\phi$ to avoid the reward from being overly weighted towards one goal or the other.

The accuracy reward depends on whether the set average accuracy, $\varepsilon^*$, was satisfied since the $n$-th sensor's last transmission, as well as on the change in the estimation error since the last transmission. We define the accuracy reward as:

\begin{equation}
r_{acc}(\overline{\varepsilon}_n) =  \begin{cases} 
\operatorname{ when }\phantom{A} \overline{\varepsilon}_n \leq \varepsilon^* \phantom{,}\operatorname{ :}\\
\phantom{-12}  \big (\frac{\overline{\varepsilon}_n}{\varepsilon^*} \big)^2 + \Upsilon \phantom{,} \Delta\overline{\varepsilon}_n\\
\operatorname{ when }\phantom{A} \overline{\varepsilon}_n > \varepsilon^* \phantom{,}\operatorname{ :}\\
\phantom{-} \big (\frac{\overline{\varepsilon}_n- \varepsilon^*}{\varepsilon^*} \big)^2  - \Upsilon  \phantom{,} \Delta\overline{\varepsilon}_n\\
\end{cases}
\label{acc_rewd}
\end{equation}

\noindent where $\Delta\overline{\varepsilon}_n$ represents the change in the average estimation error since the previous transmission. The closer the estimation error is to the target, the greater will be the reward from the first term of the expression. The second term of the accuracy reward 
steers the learning agent towards keeping the \ac{mse} as close as possible to the target $\varepsilon^*$, without exceeding it. The factor $\Upsilon$ is used to balance the contributions of the two parts of the accuracy reward.

Our energy reward exploits the relationship between the update interval and a battery-powered sensor's lifetime. The longer the time between the consecutive updates, the longer the sensors' lifetime will be. Therefore, the selected energy reward is based on how the update interval is increased or decreased, as follows:

\begin{equation}
r_{en}(E_n) =  \begin{cases} 1 - \frac{N E_n}{\sum_{i=1}^N E_n}, \qquad \qquad  \text{ if } \phantom{,} T_{\hat{n}} > T_n \\
\phantom{-AA,} 0, \qquad \qquad \qquad  \text{ if } \phantom{,}  T_{\hat{n}} = T_n \\
\frac{N E_n}{\sum_{i=1}^N E_n} - 1, \qquad  \qquad \text{ if } \phantom{,}  T_{\hat{n}} <  T_n \\
\end{cases},
\end{equation}

\noindent where $E_n$ is the sensor's available energy. If a sensor has above average available energy, the energy reward should encourage the learning agent to make sure that such a sensor updates more often, and vice-versa if a sensor has below average energy. 

In the next subsection, we present our implementation of the \ac{ddpg} algorithm.

\subsection{\acl{ann} Structure }

\begin{figure} 
	\centering
	\resizebox {0.99\columnwidth} {!} {
		\def\layersep{2cm}
\begin{tikzpicture}[shorten >=1pt,->,draw=black!50, node distance=\layersep]
    \tikzstyle{every pin edge}=[<-,shorten <=1pt]
    \tikzstyle{neuron}=[circle,fill=black!25,minimum size=17pt,inner sep=0pt]
    \tikzstyle{input neuron}=[neuron, fill=green!50];
    \tikzstyle{output neuron}=[neuron, fill=red!50];
    \tikzstyle{hidden neuron}=[neuron, fill=blue!50];
    \tikzstyle{hidden neuron_l1}=[neuron, fill=blue!75];
    \tikzstyle{annot} = [text width=4em, text centered]

    \node[input neuron, pin=left:$T_n$] (I-1) at (0,-1) {};
    \node[input neuron, pin=left:$\frac{\overline{\varepsilon}_n}{\varepsilon^*}$] (I-2) at (0,-2) {};
    \node[input neuron, pin=left:$E_n$] (I-3) at (0,-3) {};
    \node[input neuron, pin=left:$\left(\prod _{\substack{i=1 \\ i\neq n}}^{N}T_{i}\right)^{\frac {1}{N-1}}$] (I-4) at (0,-4) {};
    
    \node[input neuron, pin=left:$\left(\prod _{\substack{i=1 \\ i\neq n}}^{N}E_{i}\right)^{\frac {1}{N-1}}$] (I-5) at (0,-5) {};
    \node[input neuron, pin=left:$\left(\prod _{\substack{i=1 \\ i\neq n}}^{N} \frac{\overline{\varepsilon}_{i}}{\varepsilon^*}\right)^{\frac {1}{N-1}}$] (I-6) at (0,-6) {};
    

    \foreach \name / \y in {1,...,3}
        \path[yshift=0.5cm]
            node[hidden neuron_l1] (H1-\name) at (\layersep,-\y ) {};
    
    \foreach \name / \y in {5,...,7}
        \path[yshift=0.5cm]
            node[hidden neuron_l1] (H1-\name) at (\layersep,-\y ) {};

    \foreach \source in {1,...,6}
        \foreach \dest in {1,...,3}
            \path (I-\source) edge (H1-\dest);
    \foreach \source in {1,...,6}
        \foreach \dest in {5,...,7}
            \path (I-\source) edge (H1-\dest);

    \foreach \name / \y in {1,...,3}
        \path[yshift=0.5cm]
            node[hidden neuron] (H2-\name) at (2*\layersep,-\y ) {};
     
     \foreach \name / \y in {5,...,7}
        \path[yshift=0.5cm]
            node[hidden neuron] (H2-\name) at (2*\layersep,-\y ) {};
     \foreach \source in {1,...,3}
        \foreach \dest in {1,...,3}
            \path [black, very thick, dotted] (H1-\source) edge (H2-\dest);
    \foreach \source in {1,...,3}
        \foreach \dest in {5,...,7}
            \path [black, thick, dotted] (H1-\source) edge (H2-\dest);
    
    \foreach \source in {5,...,7}
        \foreach \dest in {1,...,3}
            \path [black, thick, dotted] (H1-\source) edge (H2-\dest);
    \foreach \source in {5,...,7}
        \foreach \dest in {5,...,7}
            \path [black, thick, dotted] (H1-\source) edge (H2-\dest);
    
    \foreach \name / \y in {1,...,3}
        \path[yshift=0.5cm]
            node[hidden neuron] (H3-\name) at (3*\layersep,-\y ) {};
     
     \foreach \name / \y in {5,...,7}
        \path[yshift=0.5cm]
            node[hidden neuron] (H3-\name) at (3*\layersep,-\y ) {};
    
     \foreach \source in {1,...,3}
        \foreach \dest in {1,...,3}
            \path [black, thick, dotted] (H2-\source) edge (H3-\dest);
    \foreach \source in {1,...,3}
        \foreach \dest in {5,...,7}
            \path [black, thick, dotted] (H2-\source) edge (H3-\dest);
    
    \foreach \source in {5,...,7}
        \foreach \dest in {1,...,3}
            \path [black, thick, dotted](H2-\source) edge (H3-\dest);
    \foreach \source in {5,...,7}
        \foreach \dest in {5,...,7}
            \path [black, thick, dotted] (H2-\source) edge (H3-\dest);
            
    \foreach \name / \y in {1,...,2}
        \path[yshift=0cm]
            node[hidden neuron] (H4-\name) at (4*\layersep,-\y ) {};
     
     \foreach \name / \y in {5,...,6}
        \path[yshift=0cm]
            node[hidden neuron] (H4-\name) at (4*\layersep,-\y ) {};
    
     \foreach \source in {1,...,3}
        \foreach \dest in {1,...,2}
            \path [black, thick, dotted] (H3-\source) edge (H4-\dest);
    \foreach \source in {1,...,3}
        \foreach \dest in {5,...,6}
            \path [black, thick, dotted] (H3-\source) edge (H4-\dest);
    
    \foreach \source in {5,...,7}
        \foreach \dest in {1,...,2}
            \path [black, thick, dotted] (H3-\source) edge (H4-\dest);
    \foreach \source in {5,...,7}
        \foreach \dest in {5,...,6}
            \path [black, thick, dotted] (H3-\source) edge (H4-\dest);
    
    \path (H1-3) -- (H1-5) node [black, font=\Huge, midway, sloped] {$\dots$};
    \path (H2-3) -- (H2-5) node [black, font=\Huge, midway, sloped] {$\dots$};
    \path (H3-3) -- (H3-5) node [black, font=\Huge, midway, sloped] {$\dots$};
    \path (H4-2) -- (H4-5) node [black, font=\Huge, midway, sloped] {$\dots$};

    
    \node[output neuron, pin={[pin edge={->}]right:$a_n$}] (O-1) at (5*\layersep,-3 - 0.5) {};
     
    
    \foreach \source in {1,...,2}
        \foreach \dest in {1,...,1}
            \path (H4-\source) edge (O-\dest);
    \foreach \source in {5,...,6}
        \foreach \dest in {1,...,1}
            \path (H4-\source) edge (O-\dest);
            


    \node[annot,above of=H2-1, node distance=1cm] (hl2) {Hidden Layer 2};
    \node[annot,left of=hl2] (hl1){Hidden Layer 1};
    \node[annot,left of=hl1] {Input};
    \node[annot,right of=hl2] (hl3){Hidden Layer 3};
    \node[annot,right of=hl3] (hl4){Hidden Layer 4};
    \node[annot,right of=hl4] {Output};
\end{tikzpicture}
	}
	\caption{Actor's ANN structure.}
	\label{fig:ann_actor}
\end{figure}
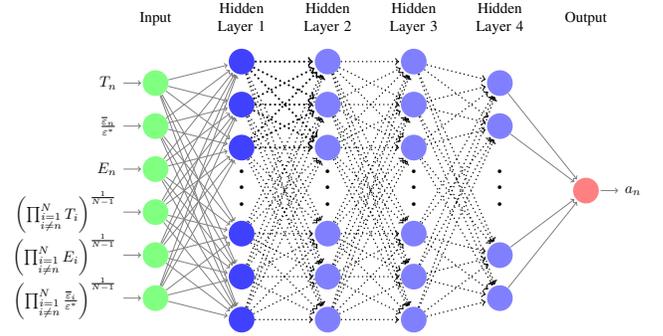

To implement the learning agent using \ac{ddpg} as described in \cite{lillicrap2015continuous} we used Pytorch \cite{paszke2017automatic}, a standard Python-based library for implementing \ac{drl} algorithms. We employ similar \ac{ann} structures for the actor and the critic. The actor's \ac{ann} consists of an input (state space), output (action value), and four hidden layers with \ac{fnn} structure, as shown in Figure \ref{fig:ann_actor}. We use 75 neurons in the first three hidden layers and 25 neurons in the fourth layer. Between each hidden layer, we implemented a $50 \%$ dropout layer. We use batch normalization after activation in the first hidden layer. The dropout layers prevent over-fitting, and batch normalization improves the learning speed. We employ the same structure for the critic's \ac{ann}, with a slight difference in the activation function used. We use a ReLU activation function for every layer in both \acp{ann}. The only exception is the output layer of the actor's \ac{ann}, where we use a Hyperbolic function. Such a difference is required as the actor's output value is limited to values between $-1$ and $1$ while the critic's is not. To train the \acp{ann} we periodically perform batch learning. In each batch training we use 128 experiences. Each experience consists of a state, corresponding action and reward, and the state to which our sensor transits after taking the selected action. Note that 128 experiences are randomly selected from a memory pool of up to 100,000 experiences. While, the exploration in \ac{ddpg} algorithms is performed by adding noise, i.e., a random value, to the actor's output value. We use the Ornstein-Uhlenbeck process~\cite{lillicrap2015continuous} to generate a random value to be added to the selected action value.

In the gateway, we deploy \textit{only one set} of actor's and critic's \ac{ann} for the \ac{ddpg} algorithm to control the update times of all $N$ sensors in the deployment. The main benefit of using only one set of \acp{ann} is the faster collection of necessary experiences for training. Additionally, the system experiences a greater variety in the six-dimensional state vector values, which means that the initial training period is much lower than if the gateway relied on a separate set of \acp{ann} for each sensor under its control.
Using only one set also has a practical benefit as it requires much less computational power at the gateway for the training process.

The \textbf{interpreter} has a crucial role in the operation of the proposed scheduler. Its function is to determine the state(the six dimension vector defined in Eq. \eqref{eq:state_space}) and reward (Eq. \eqref{eq:reward}) whenever the gateway receives an observation from the $n$-th sensor. To determine both, the interpreter relies on already gathered information regarding the environment stored in the gateway. For example, by examining metadata in the observations, the gateway can learn about the sensors' energy levels. Additionally, these observations are used to extract the covariance model's scaling parameters. The covariance model is then used to determine the estimation error. In other words, the gateway leverages collected observations to acquire knowledge about the environment to make a more informed decision. Such operational logic follows our discussion in the previous section (problem formulation).
In our design, the obtained external information is passed to the learning agent trough the interpreter's in the form of the state and reward, as illustrated in the Fig. \ref{fig:DDPG}.

We set the start of an episode to when a sensor transmits an observation. When the gateway receives an observation, it uses the interpreter to determine the sensor's state and calculates the reward for the action taken, i.e., the change in the update interval. The sensor state information is then passed to the learning agent to determine the new action; the learning agent calculates the sensor's new update interval and informs the sensor. The sensor will enter sleep mode for the amount of time determined by the learning agent. As soon as the sensor wakes up, the episode ends, and the new episode starts as the sensor transmits a new update. 

In the next section, we evaluate the performance of our proposed mechanism using data obtained from a real sensor network deployment.


\section{Evaluation and Results}
\label{sec:validation}

In this section, we evaluate our proposed scheduling mechanism using observations provided by the Intel Berkeley Research laboratory \cite{bodik2004intel}, as well as data collected from multiple sensors deployed in the city of Santander, Spain, as part of the SmartSantander testbed \cite{gutierrez2016co}. In our experiments, a simulated sensor transmits an observation with the exact value as was obtained by its real world counterpart. In other words, the data enables us to realistically represent how the observed physical phenomenon, i.e., $Z(\bm{X},t)$, varies over time. We evaluate our mechanism using five different datasets: two from Intel (temperature and humidity), and three from SmartSantander (temperature, humidity, and ambient noise). We split the evaluation into three parts: In the first part, we demonstrate that the proposed scheduling mechanism can learn the optimal behaviour and that using our approach can significantly extend the sensors' lifespan. In the second part of our evaluation, we perform complexity and run-time analysis. Finally in our third part, we highlight the energy-aware aspects of our proposed scheduling mechanism.

In our simulation, each sensor is assigned observations obtained by a real sensor and location as it had in real deployment. 
Meaning, that when a $n$-th sensor decides to transmit an observation ($\bm{X}_n$) at time $t_n$, the gateway receives a measurement, e.g., temperature reading, which the simulated sensor real counterpart has obtained in the corresponding time moment. The Intel laboratory dataset provides observations collected over nine days of measurements from 50 sensors. The observations are collected very frequently (every $31$ seconds), providing us with a ground truth of observed values for the evaluation process. On the other hand, the Smart Santander data represents a realistic deployment of sensors in a smart city environment. In our analysis, we use data obtained in the first nineteen days of April 2018. We rely on data from twenty temperature, ten humidity, and eight ambient noise sensors. In contrast to the Intel data, every sensor transmitted observations at different time intervals, and sometimes there were a few hours during which a sensor did not send any observation. We use the Amelia II software program \cite{amelia}, a tool to generate missing data, to produce the missing values. To generate these values, we used observations from the sensor that collected the highest number of observations; as a consequence, we had to remove that sensor from each SmartSantander dataset, to avoid adding bias in the evaluation. 

Due to our use of real observations to model the environment, we had to insert an extra step at the evaluation stage. The learning agent could recognise patterns in the collected observations and adapt its behaviour accordingly. In other words, the agent could overfit its policy to the dataset. To avoid such a scenario, we have to ensure that the environment the agent interacts with during evaluation is new. To that end, we had to split each dataset into two parts. The first part is used for exploration, during which the agent learns the policy. The second part is used for the evaluation of its performance.  We split the Intel dataset into six days used for exploration and three days for evaluation, while for SmartSantander we use the last six days for evaluation and the rest to train the agent.

\begin{table}[ht]
	\centering
	\caption{Static Simulation Parameters}
	\label{table1}
	\begin{tabular}{p{1.5cm} p{1.5cm}|| p{1.5cm} p{1.5cm}}
		\toprule
		\begin{tabular}[c]{@{}c@{}} Parameter \end{tabular} & 
		\begin{tabular}[c]{@{}c@{}} Value \end{tabular} & 
		\begin{tabular}[c]{@{}c@{}} Parameter \end{tabular} & 
		\begin{tabular}[c]{@{}c@{}} Value \end{tabular} \\ 
		\arrayrulecolor{black}\hline 
		\midrule 
		
		\rowcolor{Gray}
		\begin{tabular}[c]{@{}c@{}} $U_{max}$ \end{tabular}  &   
		\begin{tabular}[c]{@{}c@{}}$250s$ \end{tabular} & 
		\begin{tabular}[c]{@{}c@{}}$P_c$\end{tabular} & 
		\begin{tabular}[c]{@{}c@{}}$15uW$\end{tabular} \\

		\begin{tabular}[c]{@{}c@{}} $E_0$ \end{tabular} & 
		\begin{tabular}[c]{@{}c@{}} $6696 J$ \end{tabular} & 
		\begin{tabular}[c]{@{}c@{}} $E_{tr}$  \end{tabular}  & 
		\begin{tabular}[c]{@{}c@{}}  $78.7mJ$ \end{tabular}  \\
		
		\rowcolor{Gray}
		
	    \begin{tabular}[c]{@{}c@{}} $T_{start}$ \end{tabular} & 
		\begin{tabular}[c]{@{}c@{}} $900s$ \end{tabular} & 
		\begin{tabular}[c]{@{}c@{}} $\Upsilon$ \end{tabular} & 
		\begin{tabular}[c]{@{}c@{}} 10 \end{tabular} \\	
		
		\begin{tabular}[c]{@{}c@{}}  $\varepsilon^*$ \end{tabular}  & 
		\begin{tabular}[c]{@{}c@{}}  0.01 \end{tabular}  &
		\begin{tabular}[c]{@{}c@{}} $\phi$ \end{tabular}  & 
		\begin{tabular}[c]{@{}c@{}} $0.5$  \end{tabular}  \\
		
		\rowcolor{Gray}
	    \begin{tabular}[c]{@{}c@{}} $T_{max}$  \end{tabular} & 
		\begin{tabular}[c]{@{}c@{}} $7200s$ \end{tabular}& 
		\begin{tabular}[c]{@{}c@{}}  
		$t_S$       \end{tabular} & 
		\begin{tabular}[c]{@{}c@{}} $10s$    \end{tabular} \\

		
		
		\bottomrule
	\end{tabular}
\end{table}

We list system parameters that are kept constant throughout our evaluation process in Table \ref{table1}. We selected static simulation energy parameters by assuming that each of the sensors is powered by a single non-rechargeable Lithium coin battery with a capacity of $620$ mAh, which provides us with the value for $E_0$. The selected energy consumption parameters, i.e., $P_c$ and $E_{tr}$, mimic the power consumption of an \ac{iot} sensor using a \ac{lorawan} radio. We obtain the power parameters following the analysis presented in \cite{costa2017energy}. The selected $U_{max}$ yielded the best average performance at the end of the training phase for all five datasets. Altogether our agent can select among 51 different actions, i.e., changes of update interval.

$T_{start}$ represents a suitable starting update time value, while $T_{max}$ represents the maximal value that should be allowed between two consecutive updates from one sensor. At the start of every simulation, we set the same initial update time for every sensor. $\phi$ and $\varepsilon^*$ are set to the value stated in the Table unless they are parameters that we change in the presented experiment. We perform simulations in ten-second  time-steps. In Table \ref{table_hyper_param} we list the \ac{drl} solution hyperparameters we determined through a grid search to be most suitable. Note that we multiply the calculated rewards by a factor of $10$ to improve the training process, as higher reward values tend to reduce the time required for the \ac{ddpg} algorithm to arrive at the optimal policy~\cite{sorg2010reward}.

\begin{table}[ht]
	\centering
	\caption{Hyperparameters}
	\label{table_hyper_param}
	\begin{tabular}{p{1.5cm} p{4cm} p{1.5cm}}
		\toprule
		\multicolumn{1}{l}{}                  
		\begin{tabular}[c]{@{}c@{}}  \end{tabular} & 
		\begin{tabular}[c]{@{}c@{}} Hyperparameter\end{tabular} & 
		\begin{tabular}[c]{@{}c@{}} Value \end{tabular} \\
		\arrayrulecolor{black}\hline 
		\midrule 
		
		\begin{tabular}[c]{@{}c@{}}   \end{tabular} & 
		\begin{tabular}[c]{@{}c@{}}  Learning rate  
		\end{tabular} & 
		\begin{tabular}[c]{@{}c@{}}   $0.5$ \end{tabular} \\
		
		\begin{tabular}[c]{@{}c@{}}   \end{tabular} & 
		\begin{tabular}[c]{@{}c@{}}  Discount factor  
		\end{tabular} & 
		\begin{tabular}[c]{@{}c@{}}   $0.2$ \end{tabular} \\
		
		\begin{tabular}[c]{@{}c@{}}   \end{tabular} & 
		\begin{tabular}[c]{@{}c@{}}  Explore rate  
		\end{tabular} & 
		\begin{tabular}[c]{@{}c@{}}   $0.15$ \end{tabular} \\
		
		\begin{tabular}[c]{@{}c@{}}   \end{tabular} & 
		\begin{tabular}[c]{@{}c@{}}   Target ANN soft update 
		\end{tabular} & 
		\begin{tabular}[c]{@{}c@{}}   $10^{-3}$\end{tabular} \\
		
		\begin{tabular}[c]{@{}c@{}}   DQN \end{tabular} & 
		\begin{tabular}[c]{@{}c@{}}   ANN Learning rate \end{tabular} & 
		\begin{tabular}[c]{@{}c@{}}   $10^{-3}$ \end{tabular} \\
		
		\begin{tabular}[c]{@{}c@{}}   \end{tabular} & 
		\begin{tabular}[c]{@{}c@{}}  Batch size   \end{tabular} & 
		\begin{tabular}[c]{@{}c@{}}   $32$ \end{tabular} \\
		
		\begin{tabular}[c]{@{}c@{}}   \end{tabular} & 
		\begin{tabular}[c]{@{}c@{}}  Memory  size \end{tabular} & 
		\begin{tabular}[c]{@{}c@{}}   $2 \times 10^4$ \end{tabular} \\
		
		\begin{tabular}[c]{@{}c@{}}   \end{tabular} & 
		\begin{tabular}[c]{@{}c@{}}  Optimizer  \end{tabular} & 
		\begin{tabular}[c]{@{}c@{}}   Adam \end{tabular} \\
		
		\begin{tabular}[c]{@{}c@{}}   \end{tabular} & 
		\begin{tabular}[c]{@{}c@{}}  Loss Function  \end{tabular} & 
		\begin{tabular}[c]{@{}c@{}}   MSE \end{tabular} \\
		
		\arrayrulecolor{black}\hline
		\begin{tabular}[c]{@{}c@{}}   \end{tabular} & 
		\begin{tabular}[c]{@{}c@{}}  Actor's ANN learning rate $\tau_A$ \end{tabular} & 
		\begin{tabular}[c]{@{}c@{}}   $10^{-4}$\end{tabular} \\
		
		\begin{tabular}[c]{@{}c@{}}   \end{tabular} & 
		\begin{tabular}[c]{@{}c@{}}  Critic's ANN learning rate $\tau_C$ \end{tabular} & 
		\begin{tabular}[c]{@{}c@{}}   $10^{-4}$\end{tabular} \\
		
		\begin{tabular}[c]{@{}c@{}}   \end{tabular} & 
		\begin{tabular}[c]{@{}c@{}}  Target ANN soft update  \end{tabular} & 
		\begin{tabular}[c]{@{}c@{}}   $10^{-3}$\end{tabular} \\
		
		\begin{tabular}[c]{@{}c@{}}   \end{tabular} & 
		\begin{tabular}[c]{@{}c@{}} Critic's Discount factor \end{tabular} & 
		\begin{tabular}[c]{@{}c@{}}   0.99 \end{tabular} \\

		\begin{tabular}[c]{@{}c@{}}  DDPG  \end{tabular} & 
		\begin{tabular}[c]{@{}c@{}}  Batch size  $M$ \end{tabular} & 
		\begin{tabular}[c]{@{}c@{}}   $128$ \end{tabular} \\
		
		\begin{tabular}[c]{@{}c@{}}   \end{tabular} & 
		\begin{tabular}[c]{@{}c@{}}  Memory size \end{tabular} & 
		\begin{tabular}[c]{@{}c@{}}   $10^5$\end{tabular} \\
		
		\begin{tabular}[c]{@{}c@{}}   \end{tabular} & 
		\begin{tabular}[c]{@{}c@{}}  Optimizer  \end{tabular} & 
		\begin{tabular}[c]{@{}c@{}}   Adam \end{tabular} \\
		
		\begin{tabular}[c]{@{}c@{}}   \end{tabular} & 
		\begin{tabular}[c]{@{}c@{}}  Loss Function  \end{tabular} & 
		\begin{tabular}[c]{@{}c@{}}   MSE \end{tabular} \\
		
		\bottomrule   
\end{tabular}
\end{table}

\subsection{Performance Evaluation}

In this subsection, we evaluate our scheduling mechanism's ability to maintain accurate observations and compare it to other scheduling approaches.

\begin{figure}
	\centering

\begin{tikzpicture}
\begin{groupplot}[group style={group name=my_plots, group size=2 by 1,vertical sep= 1.75cm}, width=0.45\textwidth]]





\nextgroupplot[
axis lines=middle,
title = (a) Lifetime,
xlabel=$N$,
ylabel=$\mathcal{L}$(years),
ytick ={2.5, 3, 3.5, 4, 4.5, 5},
yticklabels={2.5, 3, 3.5, 4, 4.5, 5},
xtick={10,20,30,40,50},
legend style={at={(0.6,0.85)}},
xmin= 7.5,xmax=52.5,
ymin= 2.25,ymax= 5.75,
every tick/.style={
	black,
	semithick,
}]
\draw[black!60,dashed] (0,275) -- (435, 275);
\draw[black!60,dashed] (0,225) -- (435, 225);
\draw[black!60,dashed] (0,175) -- (435, 175);
\draw[black!60,dashed] (0,125) -- (435, 125);
\draw[black!60,dashed] (0,75) -- (435, 75);
\draw[black!60,dashed] (0,25) -- (435, 25);

\addplot[red, line width=2pt, mark=diamond*] 
table[x=density, y=DDPG_hum_life, col sep=comma]{csv_data/N_results.csv};

\addplot[blue, line width=2pt, mark=*] %
table[x=density, y=DDPG_tem_life, col sep=comma]{csv_data/N_results.csv};

\addlegendentry{Intel-Humidity};
\addlegendentry{Intel-Temperature};




\nextgroupplot[
axis lines=middle,
title = (b) Lifetime Gain,
xlabel=$N$,
ylabel= $\eta$,
ytick ={30, 35, 40, 45, 50, 55, 60},
xtick={10,20,30,40,50},
legend style={at={(0.6,0.85)}},
xmin= 7.5,xmax=52.5,
ymin= 30,ymax= 65,
every tick/.style={
	black,
	semithick,
}]

\draw[black!60,dashed] (0,300) -- (435, 300);
\draw[black!60,dashed] (0,250) -- (435, 250);
\draw[black!60,dashed] (0,200) -- (435, 200);
\draw[black!60,dashed] (0,150) -- (435, 150);
\draw[black!60,dashed] (0,100) -- (435, 100);
\draw[black!60,dashed] (0,50) -- (435, 50);

\addplot[red, line width=2pt, mark=diamond*] 
table[x=density,y=DDPG_hum_ratio, col sep=comma]{csv_data/N_results.csv};

\addplot[blue, line width=2pt, mark=*] %
table[x=density,y=DDPG_tem_ratio, col sep=comma]{csv_data/N_results.csv};

\addlegendentry{Intel-Humidity};
\addlegendentry{Intel-Temperature};

\end{groupplot}
\end{tikzpicture}
	\caption{Lifetime of sensors and lifetime gain achieved by our updating mechanism as the number of sensors under its control increases.}
	\label{fig:multiple_sources}
	\vspace{-10pt}
\end{figure}
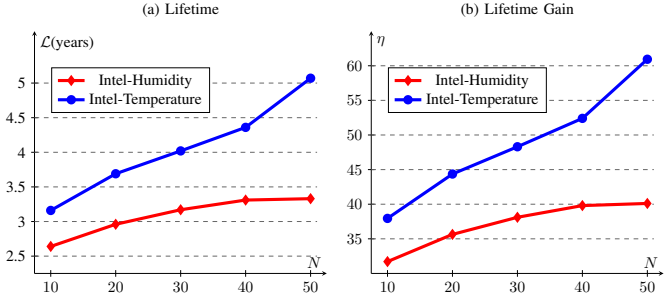

First, we test the updating mechanism performance as the number of sensors, $N$, under its management increases. The Intel datasets offer us a maximum of $50$ sensors, and for cases in which $N<50$, we randomly selected a subset of these sensors and then repeated the experiment several times. As expected, increasing the number of sensors leads to more correlated information available, and therefore, the gain of using correlated information increases with the number of sensors. The benefit of using correlated information is higher when observing temperature, due to higher correlation exhibited in the observations collected.  We calculate the expected lifetime using Eq.~\eqref{eq:lifetime}. In Fig.~\ref{fig:multiple_sources}(a) we use both Intel datasets, humidity and temperature, and calculate the expected lifetime using the average update interval that sensors achieved in the experiment. Additionally, to demonstrate the performance improvement brought by our solution, we calculate the lifetime gain $\eta$, plotted in Fig.~\ref{fig:multiple_sources}(b). We define this gain as the ratio between the lifetime achieved using our mechanism and that achieved in the original datasets. The resulting high gains arrive from the originally selected update intervals adopted in the datasets, which would lead sensors to last only a month (a time duration that coincides with the original time the sensors in the Intel lab were deployed). By adopting our approach, the same sensor deployment could last for five years, as the average update interval our \ac{drl} solution has determined is 49 minutes. 

\begin{figure*} 
	\centering
	\includegraphics[width=7in]{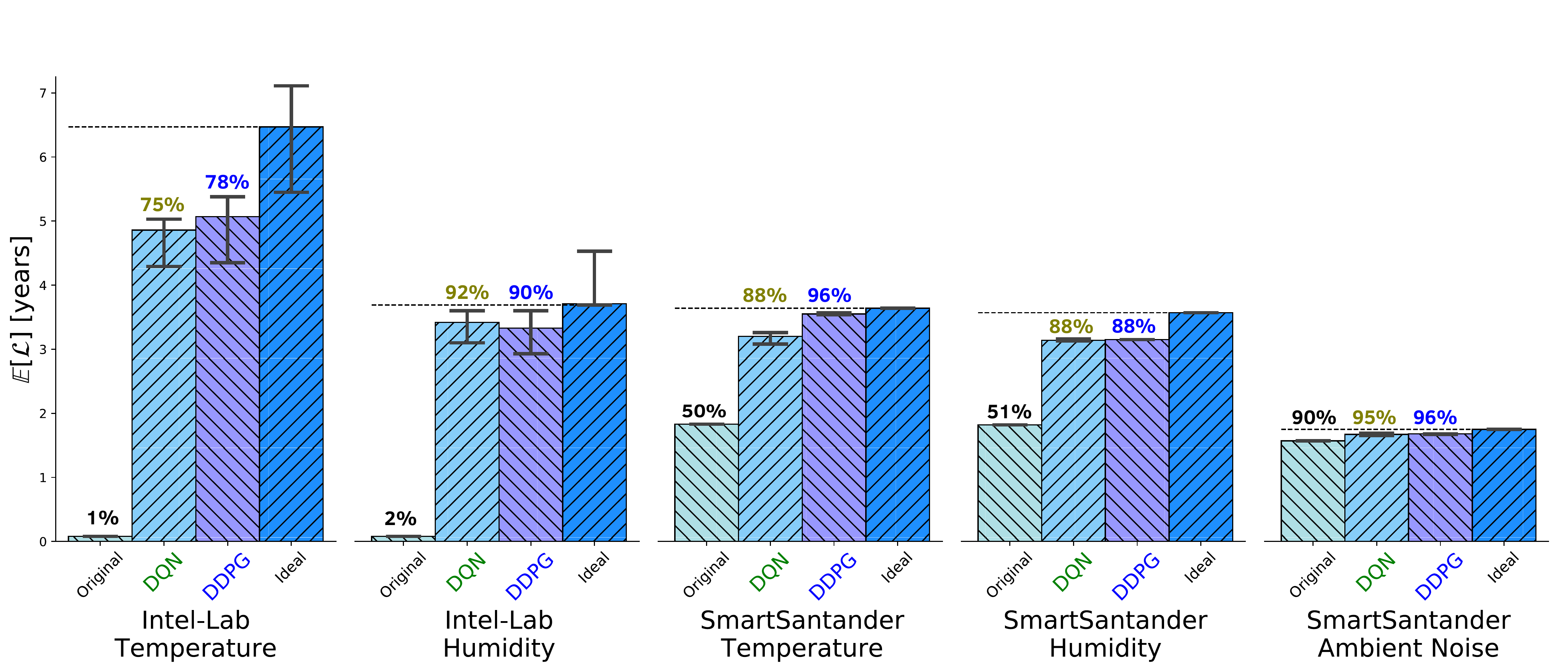}
	\caption{Achieved expected lifetime in years with every approach for all five datasets. The number on top of each bar plot reveals how close each approach is to the Ideal case.}
	\label{fig:comparison}
\end{figure*}

In Fig.~\ref{fig:comparison} we compare the performance of our \ac{ddpg}-based scheduling mechanism, in terms of achieved sensor lifetime, to three different methods of obtaining update times:
\begin{enumerate}
	\item \textbf{Original}: as a baseline, we take the update times adopted in the Intel Lab and Smart Santander datasets. The baseline case reveals the original expected lifetime of deployments, if the sensors were battery-powered. Note that for the SmartSantander datasets, we use the update time of the sensor that collected the most observations. 
	\item \textbf{\ac{dqn}}: update times we obtain by using the \acl{dqn} approach we proposed in \cite{hribar2019ICC}. Note that the reward function we use for the \ac{dqn} approach is the same as we defined in Eq. \eqref{eq:reward}, while the action space consists of five actions. The scheduling mechanism can take an action to increase the sensor's current update interval by one or ten time-steps, decrease it for one or ten time-steps, or keep it as it was.
	\item \textbf{Ideal Scheduler}: in the ideal case, the network controller acts essentially as an oracle that knows the ground truth and can obtain observations from any sensor on demand. 
	With the ideal scheduler, sensors are asked to transmit only when the average error of estimated values (of temperature, humidity, ambient noise) exceeds the set threshold $\varepsilon^*$. The lifetime the ideal scheduler obtains is the maximal possible lifetime the sensors can achieve while still maintaining the set accuracy constraint. In other words, the ideal scheduler embodies the optimal solution.
\end{enumerate}

In Fig.~\ref{fig:comparison}, we show the expected lifetime we achieved using all four approaches for all five datasets. We calculated the lifetime using Eq.~\eqref{eq:lifetime} and using values of average update times that we obtained with each approach. The results show that our mechanism is capable of finding update intervals that are very close to the ideal one. The only exception is for the case of Intel lab temperature data. In that case, a higher $U_{max}$ would enable the agent to get results closer to the ideal case. However, by keeping the $U_{max}$ constant throughout our validation we can demonstrate the system's robustness to multiple different scenarios. Similarly, for the \ac{dqn} solution, we would have to increase the number of actions to be able to improve the overall performance. However, with a higher number of actions, the required training time increases drastically, which limits the \ac{dqn} solution's applicability to the proposed problem.
Table~\ref{table_benhcmark} shows that the \ac{ddpg} approach achieves $91-92 \%$ of the error threshold,  for every dataset, while the \ac{dqn} approach varies slightly more due to the mechanism's inability to adapt to the environmental changes promptly.

The importance of faster adaptability to an ever-changing environment is paramount in real deployments. In this criterion, the \ac{ddpg} solution greatly outperforms the \ac{dqn} approach even if results in Fig.~\ref{fig:comparison} indicate very similar performance in terms of achieved lifetime. Additionally, we noticed that the \ac{dqn} approach requires a longer exploration time, e.g., for the Intel dataset, we iterated over the exploration part of the data three times, in comparison to two iterations  required to train the system using \ac{ddpg}.

To improve the performance of the proposed \ac{ddpg} approach in comparison to the ideal, the apparent solution is to provide the agent with more information, i.e., expand the \ac{ann} input state space. For example, adding information regarding the average distance to neighbouring sensors would help. However, adding the average distance as part of the state information would decrease the generality of our solution. Intuitively, providing the agent with direct information regarding all sensors in the system (current update interval, average estimation error, and energy level) should lead to better results. Unfortunately, a significantly larger state space would lead to longer computational time and even possibly a degradation in performance, as an agent could have a hard time differentiating between the more and the less relevant information, e.g., which sensor's updates are most relevant for the current decision. 

\begin{table}[ht]
	\centering
	\caption{Obtained $\overline{\varepsilon}$ per dataset.}
	\label{table_benhcmark}
		\begin{tabular}{p{2cm}|| p{1cm}| p{1cm}| p{1cm} |p{1cm}}
		\toprule
		\begin{tabular}[c]{@{}c@{}} Dataset \end{tabular} & 
		\begin{tabular}[c]{@{}c@{}} $\varepsilon^*$ \end{tabular} & 
		\begin{tabular}[c]{@{}c@{}} DQN \end{tabular} & 
		\begin{tabular}[c]{@{}c@{}} DDPG \end{tabular} &
		\begin{tabular}[c]{@{}c@{}} Ideal \end{tabular} \\
		\arrayrulecolor{black}\hline 
		\midrule
		\begin{tabular}[c]{@{}c@{}} Intel-Tem \end{tabular} & 
		\begin{tabular}[c]{@{}c@{}} $0.01$   \end{tabular} & 
		\begin{tabular}[c]{@{}c@{}} 0.0088 \end{tabular} &
		\begin{tabular}[c]{@{}c@{}} 0.0092 \end{tabular} &
		\begin{tabular}[c]{@{}c@{}} $0.01$   \end{tabular}\\
		
		\rowcolor{Gray}
		\begin{tabular}[c]{@{}c@{}} Intel-Hum \end{tabular} & 
		\begin{tabular}[c]{@{}c@{}} $0.01$   \end{tabular} & 
		\begin{tabular}[c]{@{}c@{}} 0.0093 \end{tabular} &
		\begin{tabular}[c]{@{}c@{}} 0.0091 \end{tabular} &
		\begin{tabular}[c]{@{}c@{}} $0.01$   \end{tabular}\\
		
		\begin{tabular}[c]{@{}c@{}} SmartSan-Tem  \end{tabular} & 
		\begin{tabular}[c]{@{}c@{}} $0.01$   \end{tabular} & 
		\begin{tabular}[c]{@{}c@{}} 0.0091 \end{tabular} &
		\begin{tabular}[c]{@{}c@{}} 0.0091 \end{tabular} &
		\begin{tabular}[c]{@{}c@{}} $0.01$   \end{tabular}\\
		\rowcolor{Gray}
		\begin{tabular}[c]{@{}c@{}} SmartSan-Hum  \end{tabular} & 
		\begin{tabular}[c]{@{}c@{}} $0.01$   \end{tabular} & 
		\begin{tabular}[c]{@{}c@{}} 0.0094 \end{tabular} &
		\begin{tabular}[c]{@{}c@{}} 0.0091 \end{tabular} &
		\begin{tabular}[c]{@{}c@{}} $0.01$   \end{tabular}\\
		
		\begin{tabular}[c]{@{}c@{}} SmartSan-Amb \end{tabular} & 
		\begin{tabular}[c]{@{}c@{}} $0.01$   \end{tabular} & 
		\begin{tabular}[c]{@{}c@{}} 0.0098 \end{tabular} &
		\begin{tabular}[c]{@{}c@{}} 0.0092 \end{tabular} &
		\begin{tabular}[c]{@{}c@{}} $0.01$   \end{tabular}\\
		
		\bottomrule

	\end{tabular}
\end{table}

\begin{figure}
	\centering
\begin{tikzpicture}
\begin{axis}[
axis lines=middle,
xlabel=$\overline{r}_n(m)$,
ylabel=$T_n (min)$,
ytick ={120, 150, 180, 210},
yticklabels={20, 25, 30, 35},,
xtick={2.5,5,7.5,10,12.5,15,17.5},
legend style={at={(0.95,0.615)}},
xmin=0,xmax=18,
ymin=100,ymax= 225,
every tick/.style={
	black,
	semithick,
}]
\draw[black!60,dashed] (0,110) -- (175, 110);
\draw[black!60,dashed] (0,80) -- (175, 80);
\draw[black!60,dashed] (0,50) -- (175, 50);
\draw[black!60,dashed] (0,20) -- (155, 20);

\addplot[blue, line width=2pt] 
table[x=distance,y=update_rates,col sep=comma]{csv_data/DDPG_tem_line.csv};

\addplot[
        scatter,only marks,scatter src=explicit symbolic,
        scatter/classes={
            a={mark=triangle*,blue},
            b={mark=triangle*,red},
            c={mark=o,draw=black,fill=black}
        }
    ]table[x=distance,y=update_rates, meta=label, col sep=comma]{csv_data/DDPG_tem_points.csv};


\end{axis}

\end{tikzpicture}
	\caption{Sensor's update interval change depending on the average distance to other sensors in the network.}
	\label{fig:lifetime_over_distance}
	\vspace{-10pt}
\end{figure}
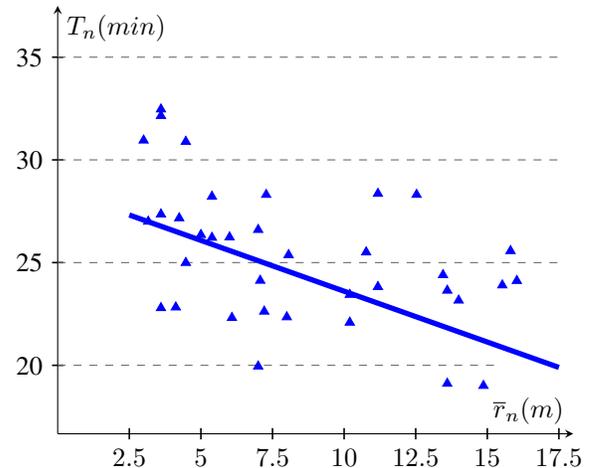

The minimum and maximum expected lifetimes among all sensors are shown as a solid line in the middle of each bar in Fig.~\ref{fig:comparison}. The variation is especially noticeable in the case of Intel-Lab data, indicating that some sensors benefit more from using correlated information than others. For this dataset, sensors are deployed on the same floor, and those in the middle of the room 
are able to benefit more from correlated measurements collected from other nearby sensors.

We remind the reader that distances are not part of the information explicitly furnished to the learning agent. As we show in Fig.~\ref{fig:lifetime_over_distance}, in which we plot the achieved average update interval as a function of the average distance to other sensors, some sensors benefit more than others from information collected by their neighbors. Note that in our system each sensor represents the location at which we observe the physical phenomenon. More specifically, it appears that the average distance to other sensors is inversely proportional to the achieved update interval. We obtained the result using temperature test data for the Intel-lab dataset. The line in Fig.~\ref{fig:lifetime_over_distance} is obtained using linear regression. Such behaviour is interesting as the agent does not have direct information regarding the distances, yet it can infer it from other available information. However, these differences appear only in the short term, as the energy-awareness ability of our proposed scheduling mechanism will in the longer run prevent one sensor from updating much more often than others and avoid the depletion of one sensor's battery much earlier than the others. 

\subsection{Complexity and Run-time analysis}

The computational complexity of our \ac{drl} solution depends solely on the dimension of the action space, $D_a$, the dimension of the state space, $D_s$, number of layers $L$, and the number of neurons in each hidden layer, $W$. Each invocation of forward propagation, i.e., when the agent makes its decision, is linear with respect to the \ac{ann} parameters. This means that we can write the number of computations required for the actor as $\mathcal{O}(D_sLW)$, and, similarly, as $\mathcal{O}((D_a+D_s)LW)$ for the critic. Considering that we employ \acp{ann} with only four hidden layers, i.e., L=4, and with 75 neurons (25 in the fourth hidden layer) per layer, the resulting number of computations is relatively small. Our agent can respond within 2 to 3 $ms$, thus satisfying the timing constraints set by the sensors' communications technology. For comparison, the \ac{dqn} solution as implemented consist of only one \ac{ann}, with an input dimension $D_{sQ}$ (state-space), number of layers, $L_Q$, number of neurons in hidden layers, $W_Q$, and actions space $D_{aQ}$. Note that the \ac{dqn} implementation in \cite{hribar2019ICC}, used in our comparison, uses a \ac{ann} with two hidden layers, each with $24$ neurons. 

In Table \ref{table_complexity}, we list computations required for both \ac{drl} solutions and on average response time. \ac{ddpg} requires more time due to the higher number of computation. However, the computation time is well within the time the gateway has at is disposal to respond.  
For comparisons, the same system would need 7-12 seconds for $N=10$  to numerically resolve the optimisation problem. Furthermore, while for a numerical solution to the optimisation problem, the complexity rises with the number of sensors to the power of four, in our problem.

\begin{table}[ht]
	\centering
	\caption{Comparison of Complexity, Implementation Run-time}
	\label{table_complexity}
	\begin{tabular}{lll}
		\toprule
		 &
		 \multicolumn{1}{c}{Computational Complexity} & 
		 \multicolumn{1}{c}{Average response time} \\
		\arrayrulecolor{black}\hline 
		\midrule 
		
         \multicolumn{1}{c}{DQN} & 
         \multicolumn{1}{c}{$\mathcal{O}(D_{sQ}  L_Q W_Q)$} & 
         \multicolumn{1}{c}{$\sim$ 2 $ms$}  \\
        
        \rowcolor{Gray}
        
         \multicolumn{1}{c}{DDPG} &  
         \multicolumn{1}{c}{$\mathcal{O}(D_sLW)$ + $\mathcal{O}((D_a+D_s)LW)$ } & 
         \multicolumn{1}{c}{2-3 $ms$} \\
        \bottomrule

\end{tabular}
\end{table}

The goal of each \ac{drl} solution is to find an updating strategy capable of prolonging the lifetime of battery-powered sensors. Experimenting with the Intel Lab dataset, we estimate that a \ac{ddpg} based scheduler achieves $80 \%$ performance of the fully converged solution. After the initial fast learning period, the performance slowly improves over the next few days due to on-line learning. In comparison, a \ac{dqn}-based scheduler requires almost three days to achieve a similar performance a \ac{ddpg}-based solution learns in one day. Furthermore, the \ac{dqn} solution requires almost twice as many days to converge fully.  Such behaviour indicates that the \ac{ddpg} implementation adapts faster to a changing environment.

\subsection{Energy-aware Scheduling}
In this subsection, we demonstrate the energy-aware capabilities of the proposed mechanism. We evaluate the energy-aware performance using Intel-laboratory temperature data, and in all experiments the scheduling mechanism controls 50 sensors.

\begin{figure}
	\centering
\begin{tikzpicture}
\begin{groupplot}[group style={group name=my_plots, group size=2 by 2,vertical sep= 1.25cm, horizontal sep= 1.35cm}, width=0.3\textwidth]
\nextgroupplot[
axis lines=middle,
xlabel=episodes,
ylabel= $T_1$(min),
xmax=350,ymax=325,
ymin=0,
xmin=0,
title = (a) $T$ for \ac{iot} Sensor 1,
legend style={at={(0.95, 0.95)}},
ytick ={60, 120, 180, 240},
yticklabels={10, 20, 30, 40},
]
\draw[black!60,dashed] (0,240) -- (350, 240);
\draw[black!60,dashed] (0,180) -- (350, 180);
\draw[black!60,dashed] (0,120) -- (350, 120);
\draw[black!60,dashed] (0,60) -- (350, 60);

\addplot[line width=1pt,solid,color=blue]%
table[x=step,y=update_rates,col sep=comma]{csv_data/sensor_low_energy.csv};

\addplot[line width=2pt,dashed,color=red]%
table[x=step,y=update1,col sep=comma]{csv_data/update_time.csv};



\addlegendentry{$T$};
\addlegendentry{$T_{avg}$};

\nextgroupplot[
axis lines=middle,
xlabel=episodes,
ylabel= $T_2$(min),
xmax=200,ymax=325,
ymin=0,
xmin=0,
title = (b) $T$ for \ac{iot} Sensor 2,
legend style={at={(0.6,0.35)}},
ytick ={60, 120, 180, 240},
yticklabels={10, 20, 30, 40},
]
\draw[black!60,dashed] (0,240) -- (500, 240);
\draw[black!60,dashed] (0,180) -- (500, 180);
\draw[black!60,dashed] (0,120) -- (500, 120);
\draw[black!60,dashed] (0,60) -- (500, 60);
\addplot[line width=1pt,solid,color=blue]%
table[x=step,y=update_rates,col sep=comma]{csv_data/sensor_high_energy.csv};

\addplot[line width=2pt,dashed,color=red]%
table[x=step,y=update2,col sep=comma]{csv_data/update_time.csv};


%
\addlegendentry{$T$};
\addlegendentry{$T_{avg}$};

\nextgroupplot[
axis lines=middle,
xlabel=episodes,
ylabel= $\overline{\varepsilon}_1(\%)$,
xmax=350,ymax=0.0125,
ymin=0,
xmin=0,
title = (c) $\overline{\varepsilon}$ for IoT sensor 1,
legend style={at={(0.6,0.35)}},
ytick ={0.0025, 0.005, 0.0075, 0.01},
yticklabels={0.0025, 0.005, 0.0075, 0.01},
ytick scale label code/.code={}
]
\draw[black!60,dashed] (0,75) -- (700, 75);
\draw[black!60,dashed] (0,25) -- (700, 25);
\draw[black!60,dashed] (0,100) -- (700, 100);
\draw[black!60,dashed] (0,50) -- (700, 50);

\addplot[line width=1pt,solid,color=blue]%
table[x=step,y=estimation_errors,col sep=comma]{csv_data/sensor_low_energy.csv};

\addplot[line width=2pt,dashed,color=red]%
table[x=step,y=error,col sep=comma]{csv_data/error_line_learn.csv};


\addlegendentry{$\overline{\varepsilon}$};
\addlegendentry{ $\varepsilon^*$};

\nextgroupplot[
axis lines=middle,
xlabel=episodes,
ylabel= $\overline{\varepsilon}_2$,
xmax=200,ymax=0.0125,
ymin=0,
xmin=0,
title = (d) $\overline{\varepsilon}$ for IoT sensor 2,
legend style={at={(0.6,0.35)}},
ytick ={0.0025, 0.005, 0.0075, 0.01},
yticklabels={0.0025, 0.005, 0.0075, 0.01},
ytick scale label code/.code={}
]
\draw[black!60,dashed] (0,75) -- (700, 75);
\draw[black!60,dashed] (0,25) -- (700, 25);
\draw[black!60,dashed] (0,100) -- (700, 100);
\draw[black!60,dashed] (0,50) -- (700, 50);

\addplot[line width=1pt,solid,color=blue]%
table[x=step,y=estimation_errors,col sep=comma]{csv_data/sensor_high_energy.csv};

\addplot[line width=2pt,dashed,color=red]%
table[x=step,y=error,col sep=comma]{csv_data/error_line_learn.csv};


\addlegendentry{$\overline{\varepsilon}$};
\addlegendentry{ $\varepsilon^*$};

\end{groupplot}

\end{tikzpicture}
	\caption{Two IoT sensors with different battery levels learning  over a number of episodes. Sensor $1$'s battery level is at $75\%$ while sensor $2$ has $25\%$ of its battery life remaining.}
	\label{fig:learning_graphs}
	\vspace{-10pt}
\end{figure}
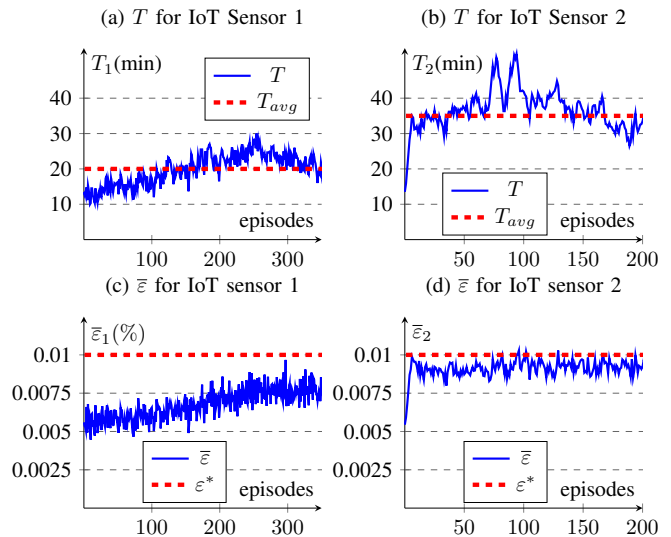

First, we show the changes in the update interval and $\overline{\varepsilon}_n$ over a number of episodes for two sensors. In Fig.~\ref{fig:learning_graphs}(a) we plot the update interval over a number of episodes for a sensor with above-average available energy ($75\%$ of remaining battery life), while in Fig.~\ref{fig:learning_graphs}(b) we plot update intervals of sensors with below-average energy ($25\%$ of battery life). Note that we set the energy levels for all other sensors to $50\%$, and that the two sensors were randomly selected. As we show, our updating mechanism sets the update interval of a sensor with less remaining energy significantly higher in comparison to the update interval of a sensor with more energy available. By setting different update intervals, the mechanism can, in the longer run, balance the energy levels among sensors. 
In other words, the mechanism forces the sensor with more energy to update more often than it would otherwise need to, in order to help preserve the energy of the sensor with shorter battery life.
As we show in Fig.~\ref{fig:learning_graphs} (c) and (d), the set error threshold is not exceeded.

\begin{figure}
	\centering
	\begin{tikzpicture}
\begin{axis}[
	ymode=log,
	xlabel=$\overline{E}_{10\%} (\%)$, 
	ylabel=$\frac{\overline{\lambda}_{10\%}}{\overline{\lambda}_{90\%}}$,
	grid=both,
	minor grid style={gray!35},
	major grid style={gray!35},
	ytick ={0.1, 0.25, 0.5, 1, 3,  5, 10, 25},
    yticklabels={0.1,0.25,0.5, 1, 3,  5, 10, 25},
	xtick ={1, 25, 50, 75, 100},
    xticklabels={1,25, 50, 75, 100},
	width=1\linewidth,
	height=0.65\linewidth,
	legend style={at={(0.4,0.95)}},
	xmin=0,xmax=105,
	ymin=0.1,ymax=30,
	every tick/.style={
	        black,
	        semithick,
	}]
\addplot[line width=1pt,solid,color=blue!75, mark=otimes*]%
	table[x=energy,y=75,col sep=comma]{csv_data/energy_ratio.csv};
\addplot[line width=1pt,solid,color=blue!50, mark=triangle] %
	table[x=energy,y=50,col sep=comma]{csv_data/energy_ratio.csv};
\addplot[line width=1pt,solid,color=blue!25, mark=square*] %
	table[x=energy,y=25,col sep=comma]{csv_data/energy_ratio.csv};

\addlegendentry{$\overline{E}_{90\%}=75\%$};
\addlegendentry{$\overline{E}_{90\%}=50\%$};
\addlegendentry{$\overline{E}_{90\%}=25\%$};
\end{axis}

\end{tikzpicture}
	\caption{The change in update rate ratio as the percentage of the battery in sensors change.}
	\label{fig:energy_savings}
	\vspace{-10pt}
\end{figure}
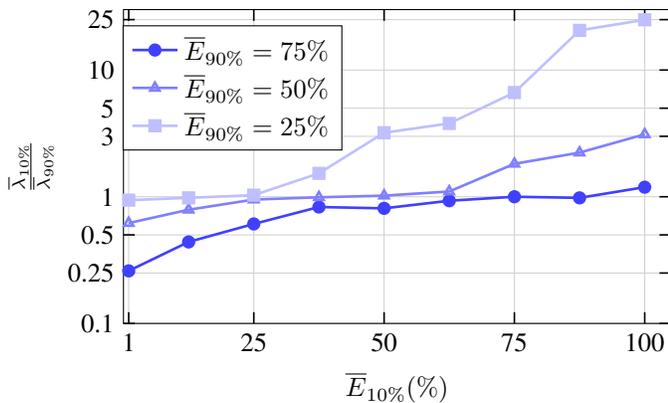

Fig.~\ref{fig:energy_savings} shows  the ratios of update rates of sensors with different energy levels. We set the energy level of 90 percent of sensors (45 out of 50) in the dataset to a fixed value. Then we change the energy level of the remaining ten percent of sensors (5) in steps from $1\%$ to $100\%$ of remaining battery life. In each step we iterate over the test data and report the ratio between the achieved average update rate of the two groups, i.e., $\frac{\overline{\lambda}_{10\%}}{\overline{\lambda}_{90\%}}$. As expected, when sensors have the same energy level, they will transmit with roughly the same update rate. We can observe an intriguing behaviour when sensors are close to depleting their energy. When a few sensors have much more energy than others, they will transmit new observations much more often, even 25 times more often when they have a full battery and the other sensors' energy level is at $25\%$. By doing so, the mechanism effectively decides to use the energy of a few sensors to prolong the lifetime of all others. 
Over time, the tendency will be towards achieving more uniform energy use throughout the network.

\section{Conclusion}
\label{sec:conclusion}

In this paper, we have proposed a \ac{drl}-based energy-aware scheduling mechanism capable of significantly prolonging the lifetime of a network of battery-powered sensors without hindering the overall performance of the sensing process. We have demonstrated, using real-world observations, that the performance of our proposed mechanism is near-optimal. Additionally, the proposed mechanism is capable of setting update intervals depending on the energy available on devices. Such behavior ensures that battery-powered sensors deployed at the same time will also expire at the same time, thus enabling infrastructure providers to replace the entire sensor deployment simultaneously. As such, the energy-aware scheduling mechanism we proposed in this paper can have a profound impact on the suitability of the future deployments of \ac{iot} sensing devices.

In our future work, we will focus on a network of sensors using diverse primary power sources, e.g., mains powered or event-based energy harvesting. In such a case, the resulting scheduling policy will depend on the type of power source the low-power sensor relies on.



\balance

\bibliographystyle{./templates/IEEEtran}
\bibliography{IEEEabrv,bibliography}

\begin{thebibliography}{10}
\providecommand{\url}[1]{#1}
\csname url@samestyle\endcsname
\providecommand{\newblock}{\relax}
\providecommand{\bibinfo}[2]{#2}
\providecommand{\BIBentrySTDinterwordspacing}{\spaceskip=0pt\relax}
\providecommand{\BIBentryALTinterwordstretchfactor}{4}
\providecommand{\BIBentryALTinterwordspacing}{\spaceskip=\fontdimen2\font plus
\BIBentryALTinterwordstretchfactor\fontdimen3\font minus
  \fontdimen4\font\relax}
\providecommand{\BIBforeignlanguage}[2]{{%
\expandafter\ifx\csname l@#1\endcsname\relax
\typeout{** WARNING: IEEEtran.bst: No hyphenation pattern has been}%
\typeout{** loaded for the language `#1'. Using the pattern for}%
\typeout{** the default language instead.}%
\else
\language=\csname l@#1\endcsname
\fi
#2}}
\providecommand{\BIBdecl}{\relax}
\BIBdecl

\bibitem{zanella2014internet}
A.~Zanella, N.~Bui, A.~Castellani, L.~Vangelista, and M.~Zorzi, ``{Internet of
  Things for Smart Cities},'' \emph{IEEE Internet Things J.}, vol.~1, no.~1,
  pp. 22--32, Feb. 2014.

\bibitem{da2014internet}
L.~Da~Xu, W.~He, and S.~Li, ``{Internet of Things in Industries: A Survey},''
  \emph{IEEE Trans. Ind. Informat.}, vol.~10, no.~4, pp. 2233--2243, Jan. 2014.

\bibitem{wolfert2017big}
S.~Wolfert, L.~Ge, C.~Verdouw, and M.-J. Bogaardt, ``{Big Data in Smart Farming
  – A review},'' \emph{Agricultural Systems}, vol. 153, pp. 69--80, May.
  2017.

\bibitem{hribar2018using}
J.~Hribar, M.~Costa, N.~Kaminski, and L.~A. DaSilva, ``{Using Correlated
  Information to Extend Device Lifetime},'' \emph{IEEE Internet Things J.},
  vol.~6, no.~2, pp. 2439--2448, Apr. 2019.

\bibitem{kosta2017age}
A.~Kosta, N.~Pappas, V.~Angelakis \emph{et~al.}, ``{Age of Information: A New
  Concept, Metric, and Tool},'' \emph{Found. Trends Netw.}, vol.~12, no.~3, pp.
  162--259, Nov. 2017.

\bibitem{kaul2011piggybacking}
S.~Kaul, R.~Yates, and M.~Gruteser, ``{On Piggybacking in Vehicular
  Networks},'' in \emph{Proc. IEEE GLOBECOM}.\hskip 1em plus 0.5em minus
  0.4em\relax Houston, TX, USA, Dec. 2011, pp. 1--5.

\bibitem{yates2015lazy}
R.~D. Yates, ``{Lazy is Timely: Status Updates by an Energy Harvesting
  Source},'' in \emph{Proc. ISIT}.\hskip 1em plus 0.5em minus 0.4em\relax Hong
  Kong, Jun. 2015, pp. 3008--3012.

\bibitem{kalor2019minimizing}
A.~E. Kalor and P.~Popovski, ``{Minimizing the Age of Information from Sensors
  with Common Observations},'' \emph{IEEE Wireless Commun. Letters}, Oct. 2019.

\bibitem{jiang2019status}
Z.~Jiang and S.~Zhou, ``{Status from a Random Field: How Densely Should One
  Update?}'' in \emph{Proc. ISIT}.\hskip 1em plus 0.5em minus 0.4em\relax
  Paris, France, Jul. 2019, pp. 1037--1041.

\bibitem{anastasi2009energy}
G.~Anastasi, M.~Conti, M.~Di~Francesco, and A.~Passarella, ``{Energy
  conservation in wireless sensor networks: A survey},'' \emph{Ad hoc
  networks}, vol.~7, no.~3, pp. 537--568, May 2009.

\bibitem{rault2014energy}
T.~Rault, A.~Bouabdallah, and Y.~Challal, ``{Energy efficiency in wireless
  sensor networks: A top-down survey},'' \emph{Computer Networks}, vol.~67, pp.
  104--122, Jul. 2014.

\bibitem{villas2014spatial}
L.~A. Villas, A.~Boukerche, H.~A. De~Oliveira, R.~B. De~Araujo, and A.~A.
  Loureiro, ``A spatial correlation aware algorithm to perform efficient data
  collection in wireless sensor networks,'' \emph{Ad Hoc Networks}, vol.~12,
  pp. 69--85, Jan. 2014.

\bibitem{yetgin2017survey}
H.~Yetgin, K.~T.~K. Cheung, M.~El-Hajjar, and L.~H. Hanzo, ``{A Survey of
  Network Lifetime Maximization Techniques in Wireless Sensor Networks},''
  \emph{IEEE Commun. Surveys Tuts.}, vol.~19, no.~2, pp. 828--854, 2nd Quart.,
  2017.

\bibitem{carrano2014survey}
R.~C. Carrano, D.~Passos, L.~C. Magalhaes, and C.~V. Albuquerque, ``{Survey and
  Taxonomy of Duty Cycling Mechanisms in Wireless Sensor Networks},''
  \emph{IEEE Commun. Surveys Tuts.}, vol.~16, no.~1, pp. 181--194, 1st Quart.,
  2014.

\bibitem{luong2018applications}
N.~C. Luong, D.~T. Hoang, S.~Gong, D.~Niyato, P.~Wang, Y.-C. Liang, and D.~I.
  Kim, ``{Applications of Deep Reinforcement Learning in Communications and
  Networking: A Survey},'' \emph{IEEE Commun. Surveys Tuts.}, vol.~21, no.~4,
  pp. 3133--3174, 4th Quart., 2019.

\bibitem{li2018q}
F.~Li, K.-Y. Lam, Z.~Sheng, X.~Zhang, K.~Zhao, and L.~Wang, ``{Q-Learning-Based
  Dynamic Spectrum Access in Cognitive Industrial Internet of Things},''
  \emph{Mobile Networks Appl.}, vol.~23, no.~6, pp. 1636--1644, Dec. 2018.

\bibitem{mohammadi2018semisupervised}
M.~Mohammadi, A.~Al-Fuqaha, M.~Guizani, and J.-S. Oh, ``{Semisupervised Deep
  Reinforcement Learning in Support of IoT and Smart City Services},''
  \emph{IEEE Internet Things J.}, vol.~5, no.~2, pp. 624--635, April 2018.

\bibitem{alsheikh2014machine}
M.~A. Alsheikh, S.~Lin, D.~Niyato, and H.-P. Tan, ``{Machine learning in
  wireless sensor networks: Algorithms, strategies, and applications},''
  \emph{IEEE Commun. Surveys Tuts.}, vol.~16, no.~4, pp. 1996--2018, 4th Quart.
  2014.

\bibitem{zheng2015green}
J.~Zheng, Y.~Cai, X.~Shen, Z.~Zheng, and W.~Yang, ``{Green Energy Optimization
  in Energy Harvesting Wireless Sensor Networks},'' \emph{IEEE Commun. Mag.},
  vol.~53, no.~11, pp. 150--157, Nov. 2015.

\bibitem{aoudia2018rlman}
F.~A. Aoudia, M.~Gautier, and O.~Berder, ``{RLMan: An Energy Manager Based on
  Reinforcement Learning for Energy Harvesting Wireless Sensor Networks},''
  \emph{IEEE Trans. Green Commun. Netw.}, vol.~2, no.~2, pp. 408--417, Jun.
  2018.

\bibitem{du2018deep}
J.~Du, H.~Chen, and W.~Zhang, ``A deep learning method for data recovery in
  sensor networks using effective spatio-temporal correlation data,''
  \emph{Sensor Review}, no.~2, pp. 208--217, Mar. 2019.

\bibitem{zhu2018new}
J.~Zhu, Y.~Song, D.~Jiang, and H.~Song, ``{A New Deep-Q-Learning-Based
  Transmission Scheduling Mechanism for the Cognitive Internet of Things},''
  \emph{IEEE Internet Things J.}, vol.~5, no.~4, pp. 2375--2385, Aug. 2018.

\bibitem{ning2019deep}
Z.~Ning, P.~Dong, X.~Wang, L.~Guo, J.~J. Rodrigues, X.~Kong, J.~Huang, and
  R.~Y. Kwok, ``{Deep Reinforcement Learning for Intelligent Internet of
  Vehicles: An Energy-Efficient Computational Offloading Scheme},'' \emph{IEEE
  Trans. Cog. Commun. Netw.}, vol.~5, no.~4, pp. 1060--1072, Dec. 2019.

\bibitem{sharma2019distributed}
M.~K. {Sharma}, A.~{Zappone}, M.~{Assaad}, M.~{Debbah}, and S.~{Vassilaras},
  ``{Distributed Power Control for Large Energy Harvesting Networks: A
  Multi-Agent Deep Reinforcement Learning Approach},'' \emph{IEEE Trans. Cog.
  Commun. Netw.}, vol.~5, no.~4, pp. 1140--1154, Dec. 2019.

\bibitem{hribar2019ICC}
J.~Hribar, A.~Marinescu, G.~A. Ropokis, and L.~A. DaSilva, ``{Using Deep
  Q-learning To Prolong the Lifetime of Correlated Internet of Things
  Devices},'' in \emph{Proc. IEEE ICC Workshops}.\hskip 1em plus 0.5em minus
  0.4em\relax Shanghai, China, May 2019, pp. 1--6.

\bibitem{bodik2004intel}
\BIBentryALTinterwordspacing
P.~Bodik, W.~Hong, C.~Guestrin, S.~Madden, M.~Paskin, and R.~Thibaux, ``{Intel
  Lab Data},'' \emph{Online dataset}, Mar. 2004. [Online]. Available:
  \url{http://db.csail.mit.edu/labdata/labdata.htm}
\BIBentrySTDinterwordspacing

\bibitem{gutierrez2016co}
V.~Guti{\'e}rrez, E.~Theodoridis, G.~Mylonas, F.~Shi, U.~Adeel, L.~Diez,
  D.~Amaxilatis, J.~Choque, G.~Camprodom, J.~McCann \emph{et~al.},
  ``{Co-Creating the Cities of the Future},'' \emph{Sensors}, vol.~16, no.~11,
  pp. 1971--1997, Nov. 2016.

\bibitem{schizas2008consensus}
I.~D. Schizas, G.~B. Giannakis, S.~I. Roumeliotis, and A.~Ribeiro, ``{Consensus
  in Ad Hoc WSNs With Noisy Links—Part II: Distributed Estimation and
  Smoothing of Random Signals},'' \emph{IEEE Trans. Signal Proc.}, vol.~56,
  no.~4, pp. 1650--1666, Apr. 2008.

\bibitem{oppenheim2015signals}
A.~V. Oppenheim and G.~C. Verghese, \emph{{Signals, Systems and
  Inference}}.\hskip 1em plus 0.5em minus 0.4em\relax Pearson, Mar. 2015.

\bibitem{cressie1999classes}
N.~Cressie and H.-C. Huang, ``{Classes of Nonseparable, Spatio-Temporal
  Stationary Covariance Functions},'' \emph{J. Amer. Statis. Assoc.}, vol.~94,
  no. 448, pp. 1330--1339, Dec. 1999.

\bibitem{gneiting2002nonseparable}
T.~Gneiting, ``{Nonseparable, Stationary Covariance Functions for Space–Time
  Data},'' \emph{J. Amer. Statis. Assoc.}, vol.~97, no. 458, pp. 590--600, Jun.
  2002.

\bibitem{chen2005lifetime}
Y.~Chen and Q.~Zhao, ``{On the Lifetime of Wireless Sensor Networks},''
  \emph{Communications letters}, vol.~9, no.~11, pp. 976--978, Nov 2005.

\bibitem{bormann2014terminology}
C.~Bormann, M.~Ersue, and A.~Keranen, ``{Terminology For Constrained-node
  Networks},'' Internet Engineering Task Force, Tech. Rep., May 2014.

\bibitem{tsoukaneri2018group}
G.~Tsoukaneri, M.~Condoluci, T.~Mahmoodi, M.~Dohler, and M.~K. Marina, ``{Group
  Communications in Narrowband-IoT: Architecture, Procedures, and
  Evaluation},'' \emph{IEEE Internet Things J.}, vol.~5, no.~3, pp. 1539--1549,
  Jun. 2018.

\bibitem{loradatasheet}
\BIBentryALTinterwordspacing
\emph{{SX1272/73 - 860 Mhz to 1020 MHz Low Power Long Range Transceiver
  Datasheet}}, SAMTECH Corporation, Jan. 2019, revision 4. [Online]. Available:
  \url{https://www.semtech.com}
\BIBentrySTDinterwordspacing

\bibitem{sutton1998reinforcement}
R.~S. Sutton and A.~G. Barto, \emph{{Reinforcement learning: An
  introduction}}.\hskip 1em plus 0.5em minus 0.4em\relax MIT press, Oct. 2018.

\bibitem{bhatnagar2008incremental}
S.~Bhatnagar, M.~Ghavamzadeh, M.~Lee, and R.~S. Sutton, ``{Incremental Natural
  Actor-critic Algorithms},'' in \emph{Proc. NIPS}.\hskip 1em plus 0.5em minus
  0.4em\relax Vancouver, Canada, 2008, pp. 105--112.

\bibitem{ddpg}
D.~Silver, G.~Lever, N.~Heess, T.~Degris, D.~Wierstra, and M.~Riedmiller,
  ``{Deterministic Policy Gradient Algorithms},'' in \emph{Proc. ICML}.\hskip
  1em plus 0.5em minus 0.4em\relax Beijing, China, Jun. 2014, pp. 1--9.

\bibitem{lillicrap2015continuous}
T.~P. Lillicrap, J.~J. Hunt, A.~Pritzel, N.~Heess, T.~Erez, Y.~Tassa,
  D.~Silver, and D.~Wierstra, ``{Continuous Control With Deep Reinforcement
  Learning},'' \emph{arXiv preprint arXiv:1509.02971}, Sep. 2015.

\bibitem{mnih2013playing}
V.~Mnih, K.~Kavukcuoglu, D.~Silver, A.~Graves, I.~Antonoglou, D.~Wierstra, and
  M.~Riedmiller, ``{Playing Atari With Deep Reinforcement Learning},''
  \emph{arXiv preprint arXiv:1312.5602}, Dec. 2013.

\bibitem{paszke2017automatic}
A.~Paszke, S.~Gross, S.~Chintala, G.~Chanan, E.~Yang, Z.~DeVito, Z.~Lin,
  A.~Desmaison, L.~Antiga, and A.~Lerer, ``{Automatic differentiation in
  PyTorch},'' in \emph{Proc. NIPS-Workshop}.\hskip 1em plus 0.5em minus
  0.4em\relax Long Beach, CA, USA, Dec. 2017, pp. 1--4.

\bibitem{amelia}
J.~Honaker, G.~King, and M.~Blackwell, ``{Amelia II: A Program for Missing
  Data},'' \emph{J. Statistical Software}, vol.~45, no.~7, pp. 1--47, Dec.
  2011.

\bibitem{costa2017energy}
M.~Costa, T.~Farrell, and L.~Doyle, ``{On Energy Efficiency and Lifetime in Low
  Power Wide Area Network for The Internet of Things},'' in \emph{Proc. IEEE
  CSCN}.\hskip 1em plus 0.5em minus 0.4em\relax Helsinki, Finland, Sep. 2017,
  pp. 258--263.

\bibitem{sorg2010reward}
J.~Sorg, R.~L. Lewis, and S.~P. Singh, ``{Reward Design Via Online Gradient
  Ascent},'' in \emph{Proc. NIPS}.\hskip 1em plus 0.5em minus 0.4em\relax
  Vancouver, Canada, Dec. 2010, pp. 2190--2198.

\end{thebibliography}

\end{document}